\documentclass[a4paper,fleqn]{latest-cas-sc}
\usepackage[numbers,sort&compress]{natbib}
\usepackage{setspace}

\usepackage{soul}
\usepackage{pifont}
\usepackage{color, xcolor}
\sethlcolor{yellow}
\usepackage{lineno}
\usepackage[ruled]{algorithm2e}

\usepackage{graphicx} 
\usepackage{subcaption} 

\def\tsc#1{\csdef{#1}{\textsc{\lowercase{#1}}\xspace}}
\tsc{WGM}
\tsc{QE}
\tsc{EP}
\tsc{PMS}
\tsc{BEC}
\tsc{DE}
\begin{document}
\soulregister{\citeyearpar}7
\soulregister{\citeyear}7
\soulregister{\ref}7
\soulregister{\citep}7
\soulregister{\cite}7
\let\WriteBookmarks\relax
\def\floatpagepagefraction{1}
\def\textpagefraction{.001}
\shorttitle{}

\shortauthors{}

\title [mode = title]{Efficient RGB-D Scene Understanding via Multi-task Adaptive Learning and Cross-dimensional Feature Guidance}          



\author[1,5]{Guodong~Sun}
\ead{sunguodong@hbut.edu.cn} 
\credit{}

\author[1,2,5]{Junjie~Liu}
\ead{102210139@hbut.edu.cn}
\credit{}

\author[1,5]{Gaoyang~Zhang}
\ead{zhanggaoyang@hbut.edu.cn}
\credit{}

\author[4]{Bo~Wu}
\ead{wubo@sari.ac.cn} 
\credit{}

\author[1,2,3,5]{Yang~Zhang}
\cormark[1] 
\ead{yzhangcst@hbut.edu.cn} 
\credit{}

\affiliation[1]{organization={School of Mechanical Engineering},
            addressline={Hubei University of Technology}, 
            city={Wuhan},
            postcode={430068}, 
            country={China}}
\affiliation[2]{organization={Vehicle Measurement, Control and Safety Key Laboratory of Sichuan Province},
            addressline={Xihua University}, 
            city={Chengdu},
            postcode={610039}, 
            country={China}}

\affiliation[3]{organization={National Key Laboratory for Novel Software Technology},
			addressline={Nanjing University}, 
			city={Nanjing},
			postcode={210023}, 
			country={China}}

\affiliation[4]{organization={Shanghai Advanced Research Institute},
		  addressline={Chinese Academy of Sciences}, 
			city={Shanghai},
			postcode={201210}, 
			country={China}}
   
\affiliation[5]{organization={Hubei Key Laboratory of Modern Manufacturing Quality Engineering},
            addressline={Hubei University of Technology}, 
            city={Wuhan},
            postcode={430068}, 
            country={China}}


\cortext[1]{Corresponding author: Yang~Zhang}


\begin{abstract}
Scene understanding plays a critical role in enabling intelligence and autonomy in robotic systems. Traditional approaches often face challenges, including occlusions, ambiguous boundaries, and the inability to adapt attention based on task-specific requirements and sample variations. To address these limitations, this paper presents an efficient RGB-D scene understanding model that performs a range of tasks, including semantic segmentation, instance segmentation, orientation estimation, panoptic segmentation, and scene classification. The proposed model incorporates an enhanced fusion encoder, which effectively leverages redundant information from both RGB and depth inputs. For semantic segmentation, we introduce normalized focus channel layers and a context feature interaction layer, designed to mitigate issues such as shallow feature misguidance and insufficient local-global feature representation. The instance segmentation task benefits from a non-bottleneck 1D structure, which achieves superior contour representation with fewer parameters. Additionally, we propose a multi-task adaptive loss function that dynamically adjusts the learning strategy for different tasks based on scene variations. Extensive experiments on the NYUv2, SUN RGB-D, and Cityscapes datasets demonstrate that our approach outperforms existing methods in both segmentation accuracy and processing speed.
\end{abstract}



\maketitle

\section{Introduction}
\label{sec:introduction}
Scene understanding empowers robots to accurately perceive their surroundings, identify objects, and classify scenes, ultimately enhancing their decision-making capabilities. Traditional scene understanding methods typically focus on a single task, limiting robots' holistic comprehension of their environments \cite{openscene}. In contrast, multi-task learning enables mutual reinforcement and synergistic optimization across multiple tasks by sharing information and learning mechanisms \cite{kbsmultitask}. This research explores a multi-task adaptive learning approach for scene understanding, aiming to design an optimal network architecture and learning strategy, thereby providing robots with more comprehensive decision support in practical applications.

Models for scene understanding typically leverage both RGB and depth information as inputs. RGB data provides essential color and texture features for object and region identification, while depth information offers precise spatial details regarding object locations and distances. Seichter et al. \cite{emsanet} employed a dual-encoder structure to separately extract features from RGB and depth inputs. However, this approach does not adequately integrate these complementary data sources, thereby missing the potential synergistic benefits they could offer. Fischedick et al. \cite{emsaformer} utilized a single Swin Transformer v2 \cite{swinv2} to jointly extract RGB and depth information. Although this method enhances integration, it involves extensive matrix computations and memory access, which can be limiting in resource-constrained environments and lead to slower processing speeds. Consequently, there is a need for an efficient encoder architecture that can effectively extract complementary information while maintaining a balance between processing speed and accuracy.

As a dense prediction task, panoptic segmentation integrates both semantic and instance segmentation. Park et al. \cite{bottleneck} employed a bottleneck feature extraction module, which is structurally simple and requires fewer parameters. However, this approach reduces feature diversity due to dimensionality reduction, thereby limiting the network’s non-linear capacity. In contrast, the non-bottleneck 1D module \cite{nonbottle1d} enhances the network's non-linearity by decomposing convolution operations, achieving a better balance between parameter efficiency and perceptual capability. The multi-layer perceptron (MLP) mechanism \cite{segformer} is also commonly used in segmentation tasks due to its robust representation abilities. However, the performance of the MLP decoder is heavily dependent on the quality of features extracted by the encoder. If the encoder fails to adequately capture critical information and details, the MLP decoder struggles to deliver accurate segmentation results \cite{shallow}. Given these considerations, effectively guiding and integrating multi-dimensional features remains a key challenge in complex scene segmentation tasks.

Multi-task learning integrates various tasks by sharing information and jointly optimizing to achieve comprehensive scene perception. However, due to the complexity and diversity of scenes, the learning difficulties, data distributions, and the relative importance of different tasks vary significantly. The fixed learning strategy often fails to adapt to these variations. Lin et al. \cite{uncertain} addressed this issue by employing random loss and stochastic gradient weighting during training, thereby avoiding human bias in weight assignment. However, the random nature of weight assignment introduces variability in model performance and can lead to instability. Liu et al. \cite{adaptive} mitigated this by calculating task weights based on training losses, which improves stability to some extent. Nevertheless, this approach adjusts weights only based on the first data batch in each iteration, lacking real-time adaptability. Therefore, a critical challenge in multi-task scene understanding is the development of a mechanism capable of adaptively adjusting weights in real time, in accordance with task characteristics and the current state of the model's training.


\begin{figure*}[!t]
	\centering
	\includegraphics[width=3.0in]{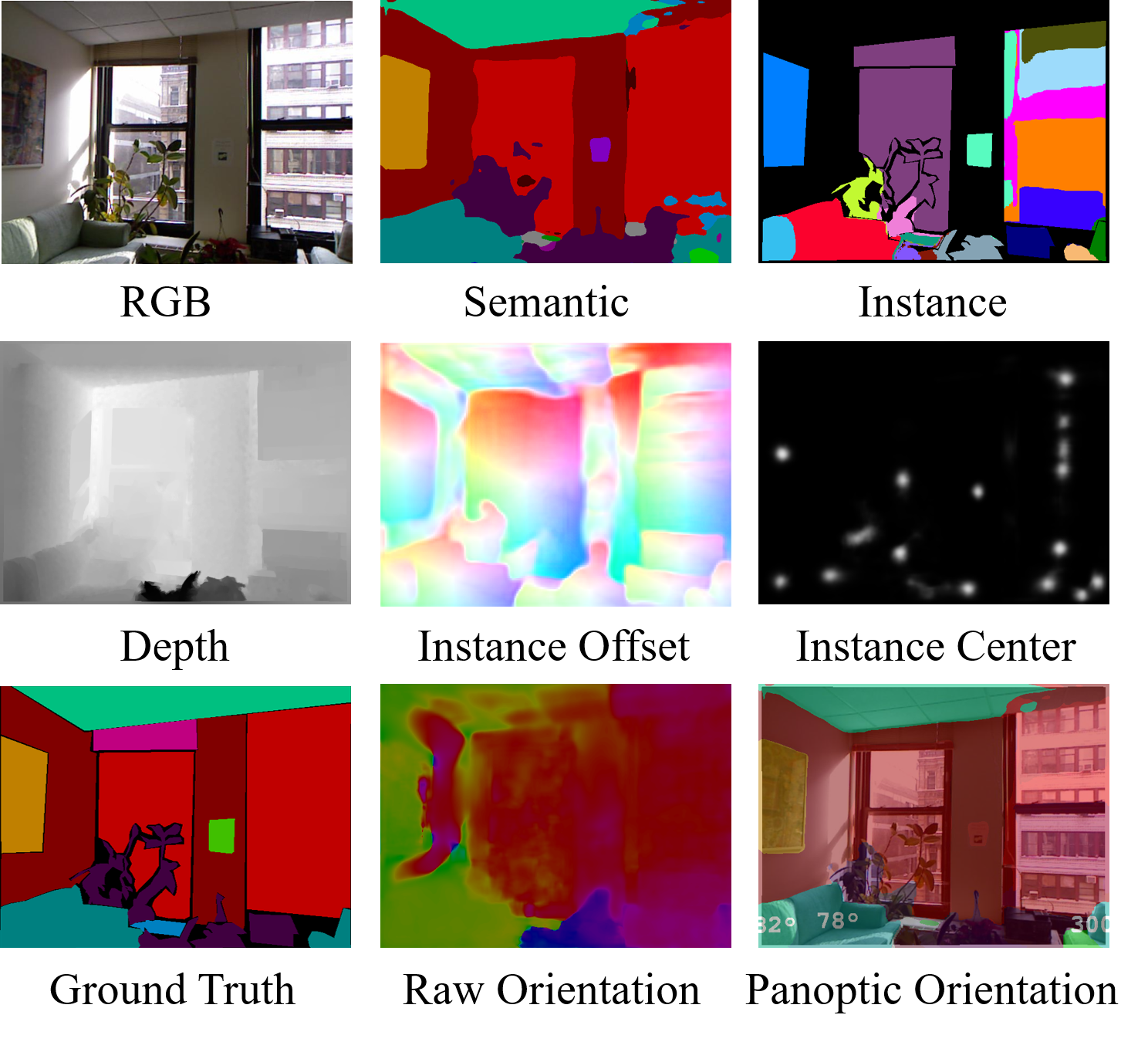}
	\caption{Multi-task outputs of the network. Semantic segmentation provides foreground masks for instance segmentation, and pixel-level instances are generated by combining instance centers with instance offsets. Panoptic segmentation is achieved by integrating semantic and instance segmentation.}
	\label{abstract1}
\end{figure*}

To address these challenges, we introduce a novel multi-task adaptive scene understanding network. The multi-task outputs of our method are illustrated in Fig. \ref{abstract1}. Our approach employs an efficient fusion encoder that simultaneously extracts complementary cues from both RGB and depth data, enabling faster processing speeds. Building on the MLP decoder \cite{segformer}, we incorporate normalized focus channel layers and a context feature interaction layer to effectively integrate spatial relationships and local information across dimensions. Furthermore, we propose a multi-task adaptive loss function to meet the varying demands of semantic segmentation, instance segmentation, orientation estimation, and scene classification tasks. This function alleviates the limitations of fixed learning strategies by dynamically adjusting learning weights based on data variations, thereby significantly improving the model's generalization performance.

The main innovations and contributions of this work include:

$\bullet $ Exploring a faster and more efficient feature extraction method that fully leverages complementary cues from both RGB and depth information.

$\bullet $ Introducing the normalized focus channel layer and context feature interaction layer to integrate local key information and spatial structure across dimensions.

$\bullet $ Designing a multi-task adaptive loss function that dynamically adjusts the learning strategy in real time based on data variations, facilitating more efficient joint training and performance optimization.

$\bullet $ Proposing a multi-task adaptive learning network for RGB-D scene understanding, enhanced with cross-dimensional feature guidance. Experiments on the NYUv2, SUN RGB-D, and Cityscapes datasets demonstrate the model's superior scene perception capabilities.

The remainder of the paper is organized as follows: Section II reviews related work in scene understanding. Section III provides a detailed description of the proposed method. Section IV presents the experiments and discussion. Finally, Section V concludes the paper.

\section{Related work}\label{sec:Related Work}
\subsection{Semantic segmentation}
Semantic segmentation aims to partition image pixels based on the objects or regions they represent, playing a critical role in various fields such as robotic perception, autonomous driving, and medical imaging. Zhao et al. \cite{pspnet} improved context aggregation through the pyramid pooling module, enabling better capture of global scene priors. However, their approach relies solely on RGB images, which provide surface-level features like color and texture but lack direct distance information. Additionally, RGB images are sensitive to changes in lighting and shadows, which can degrade performance. To enhance segmentation robustness and accuracy, there has been a growing trend toward utilizing RGB-D datasets, which combine RGB images with depth information.
The NYUv2 dataset \cite{nyuv2} is widely used for indoor scene segmentation and includes RGB images, dense depth maps, and pixel-level annotations. Similarly, SUN RGB-D \cite{sunrgbd} covers indoor scenes, offering a diverse set of object categories. The Cityscapes dataset \cite{cityscapes} specializes in urban street scenes with high-quality annotations, while ScanNet \cite{scannet} provides large-scale 3D reconstructions of indoor environments. The KITTI dataset \cite{kitti}, primarily focused on autonomous driving, is extensively used for evaluating semantic segmentation and depth estimation in outdoor environments.

Building upon these datasets, significant progress has been made in semantic segmentation methods. Qi et al. \cite{3dgnn} integrated RGB and depth data to improve the precision of spatial information. Wang et al. \cite{dcnn2018} introduced depth-aware convolution and average pooling to incorporate geometric details by considering depth similarity between pixels. Wu et al. \cite{linkrgbd} employed an interactive attention mechanism to fuse input information from dual encoders, enhancing object representation. However, while dual-encoder fusion shows promise, it introduces higher computational demands. Zhang et al. \cite{sgacnet} optimized model efficiency by utilizing depth-separable convolutions, which reduce the number of parameters and computations. Nonetheless, its higher memory access-to-computation ratio compared to standard convolutions can impact processing speed.
To achieve richer semantic representations, Wei et al. \cite{mmanet} designed an edge-aware distillation approach to guide the network to focus on boundary-adjacent samples by balancing sample contributions and classification uncertainty. Valada et al. \cite{ssma} proposed a multimodal semantic segmentation network that dynamically fuses modality features in a self-supervised manner, enabling the network to better perceive scene context, spatial positions, and object categories. Xue et al. \cite{mke} applied a knowledge distillation framework, allowing a single-modality network to perform effectively even with unlabeled multimodal data. This method effectively mitigates the high annotation cost of multimodal datasets but fails to fully leverage the complementary information across different modalities. Although these methods have advanced the field, the persistent issue of redundant modality information continues to limit model performance. Our work therefore focuses on developing a more efficient modality fusion and extraction mechanism.

\subsection{Panoptic segmentation}
In contrast to semantic segmentation, panoptic segmentation assigns pixels to semantic categories while also distinguishing individual instances within the image, providing richer information for scene understanding. Cheng et al. \cite{panoptic} pioneered this approach with a dual-decoder and atrous spatial pyramid pooling architecture, introducing the first single-step, pixel-classification-prioritized panoptic segmentation method and laying the foundation for subsequent research. However, under dynamic conditions, challenges such as motion blur and varying object sizes arise. Kachole et al. \cite{asynchronous} addressed these issues by utilizing dynamic vision sensors and a collaborative context structure to integrate multiple adjacent events, thereby generating parallel representations. To incorporate more information during the upsampling process, Fischedick et al. \cite{emsaformer} expanded the model's width and employed random masking of RGB or grayscale images during pre-training, which mitigated the reliance on a single data source. In pursuit of more robust scene understanding and cross-domain information integration, He et al. \cite{towardsdeeply} explored the correlation between depth estimation and panoptic segmentation. This approach enables cross-modal guided learning, allowing features from different modalities to mutually enhance one another. Additionally, to reduce the computational costs associated with dense operations, Cavagnero et al. \cite{pem} proposed a cross-attention algorithm that leverages visual feature redundancy to constrain computational costs without compromising model performance. The intensive pixel labeling required in panoptic segmentation increases its learning cost. To mitigate this, Li et al. \cite{point2mask} introduced a single-point supervision method, where each object is annotated with a single random point, and relatively precise pseudo-masks are generated based on category and instance differences between objects. 

These methods have achieved significant results in integrating diverse information sources and improving computational efficiency. However, due to the diversity of scenes, ensuring stable cross-domain learning and generalization capability remains challenging. Our approach further explores more effective feature guidance to extract multi-dimensional information for addressing these issues.
\subsection{Multi-task learning}
Traditionally, tasks in scene understanding have often been treated as independent, neglecting their potential intrinsic connections and interdependencies \cite{single}. With advancements in research, multi-task learning (MTL) has emerged as an effective strategy that leverages shared information and complementary characteristics across tasks, thereby enhancing overall learning efficiency. In this context, He et al. \cite{sosdnet} addressed the ambiguity in depth estimation by integrating semantic information with geometric cues and scene parsing, simultaneously improving semantic segmentation accuracy. Xu et al. \cite{padnet} employed intermediate auxiliary tasks for supervision and provided inputs for the final tasks. Moreover, Gao et al. \cite{cinet} enabled features of different tasks to guide each other by adopting a consistent learning strategy. Lopes et al. \cite{densemtl} introduced a novel bidirectional cross-task attention mechanism, combining guided cross-task attention with self-attention to strengthen the representation of task-specific features. Despite these advancements, finding the optimal balance between task-specific and shared features remains a challenging problem in MTL. Lyu et al. \cite{DualDIANet} addressed this by aggregating multi-scale, multi-level information to provide rich cues for determining which information should be shared. Liu et al. \cite{mtan} learned task-specific features from a shared network with global feature pooling to enable end-to-end training. Han et al. \cite{dmtl-net} modeled attribute correlations and heterogeneity within a single network to facilitate shared feature learning across all attributes. Previous multi-task image segmentation methods required separate training for each task to achieve optimal performance. Jain et al. \cite{oneformer} overcame this limitation by proposing a joint training strategy that learns the ground truth labels of multiple tasks in a single training process, adjusting the model using task tokens. These approaches underscore the potential of MTL to share knowledge and leverage complementary characteristics across tasks. To further adaptively adjust the learning priorities of tasks in real time, Shen et al. \cite{go4a} designed an adaptive group risk minimization strategy to explicitly optimize task learning. Agiza et al. \cite{mtlora} introduced low-rank adaptation modules to decouple the parameter space during MTL fine-tuning. Additionally, to address the issue of conflicting gradient directions in MTL, Liu et al. \cite{config} proposed a conflict-free inverse gradient method, ensuring that all loss terms are optimized at a consistent rate. 

The aforementioned methods have achieved model adaptability, but the complexity of scene data is continuously changing, and the model’s response speed remains limited. Our research aims to design a batch-level, real-time adaptive learning mechanism that dynamically adjusts the priority of task learning. This mechanism not only balances the sharing and differentiation of task features but also focuses on improving the model’s response speed and adaptability in complex environments, thereby effectively overcoming the limitations of traditional methods in delayed responses under dynamic scenarios.

\section{Method}
This section presents the architecture of the proposed model, beginning with an overview of its overall structure. Subsequently, we elaborate on the efficient feature fusion methodology and the cross-dimensional feature guidance mechanism. Finally, we introduce the multi-task adaptive learning strategy.
\label{sec:guidelines}
\subsection{Overall framework}
\begin{figure*}[!t]
	\centering
	\includegraphics[width=6.5in]{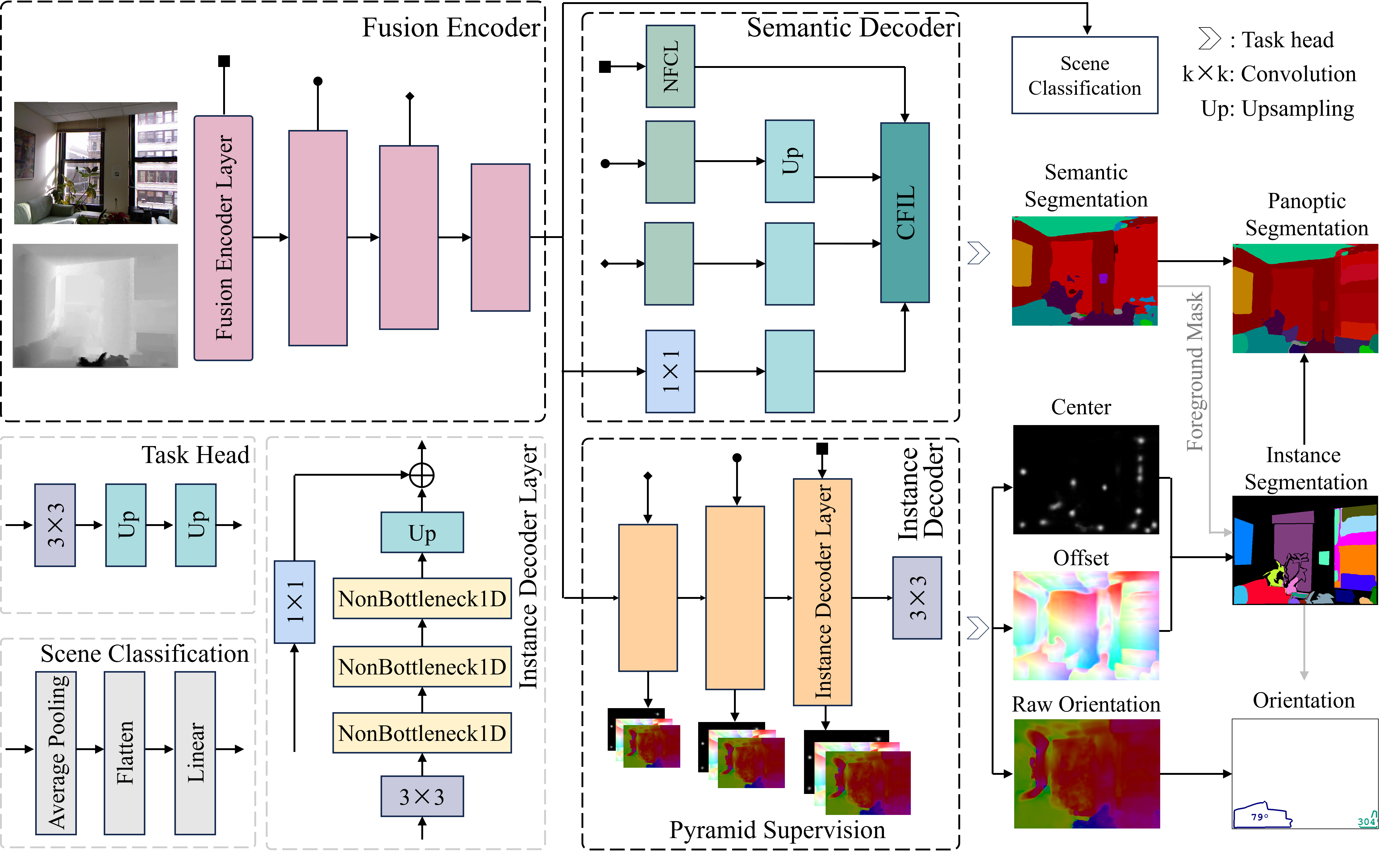}
	\caption{Multi-task scene understanding network structure. This network features an improved feature fusion encoder that handles redundant information from RGB and depth to enhance feature extraction. The semantic decoder includes NFCL and CFIL to enrich scene representations from various dimensions. The instance decoder employs the non-bottleneck 1D architecture to generate instance centers, instance offsets, and raw orientations. The integration of instance segmentation with semantic segmentation enables panoptic segmentation. Scene classification is performed by a task head with a fully connected layer. In addition, the multi-task adaptive loss function optimizes the training strategy based on data variations.}
	\label{Framework}
\end{figure*}

To achieve local and spatial perception in complex scenes and enhance generalization performance, we design a novel multi-task scene understanding network, as depicted in Fig. \ref{Framework}. This network incorporates an improved feature fusion encoder that efficiently leverages redundant features from both RGB and depth information. The extracted features are then fed into three components: a scene classification head, a semantic decoder, and an instance decoder. The scene classification task is implemented using a simple fully connected layer, while the semantic decoder is based on a lightweight multi-layer perceptron (MLP) structure. To capture key information and spatial structures, we introduce the normalized focus channel layer (NFCL) and context feature interaction layer (CFIL), which enrich scene representations across dimensions.

The instance decoder employs a lightweight non-bottleneck 1D architecture, generating instance centers, pixel offsets relative to instance centers, and raw orientations through its task head. Pyramid supervision is applied to each layer’s output in the instance decoder. Semantic segmentation provides foreground masks for instance segmentation. By combining instance centers and pixel offsets, object pixels are grouped into instance results. The combination of semantic segmentation and instance segmentation enables panoptic segmentation. The instance decoder also predicts the orientations of object instances in the scene, represented as angles around the axis perpendicular to the ground. During training, we introduce a multi-task adaptive loss function that dynamically adjusts the training strategy based on data variations.

\subsection{Efficient fusion encoder}

RGB images provide rich color information but fall short of representing geometric structures. Conversely, depth images offer accurate distance information but lack boundary details. By employing a single fusion encoder to process both RGB and depth data, we can leverage their complementary characteristics to obtain a more comprehensive scene description. Typically, scene understanding encoders are pretrained on the ImageNet \cite{imagenet} dataset to improve model accuracy. However, ImageNet contains only RGB information and lacks depth data. To address this, we sum the RGB three-channel weights as the depth weights, defined as $D=(R+G+B)/2$, avoiding the need for additional resources.

To create a more lightweight network, previous work applied depthwise separable convolution to the backbone \cite{efficientnet}. Depthwise convolution operates independently on each channel, while pointwise convolution fuses information across channels. This technique reduces the number of parameters and FLOPs. However, frequent memory accesses by these operators can impact computational speed. Given the similarity among different channel features, effectively leveraging redundant features among channels is crucial for optimizing feature extraction \cite{fasternet}.

\begin{figure}[!t]
	\centering
	\includegraphics[width=3.5in]{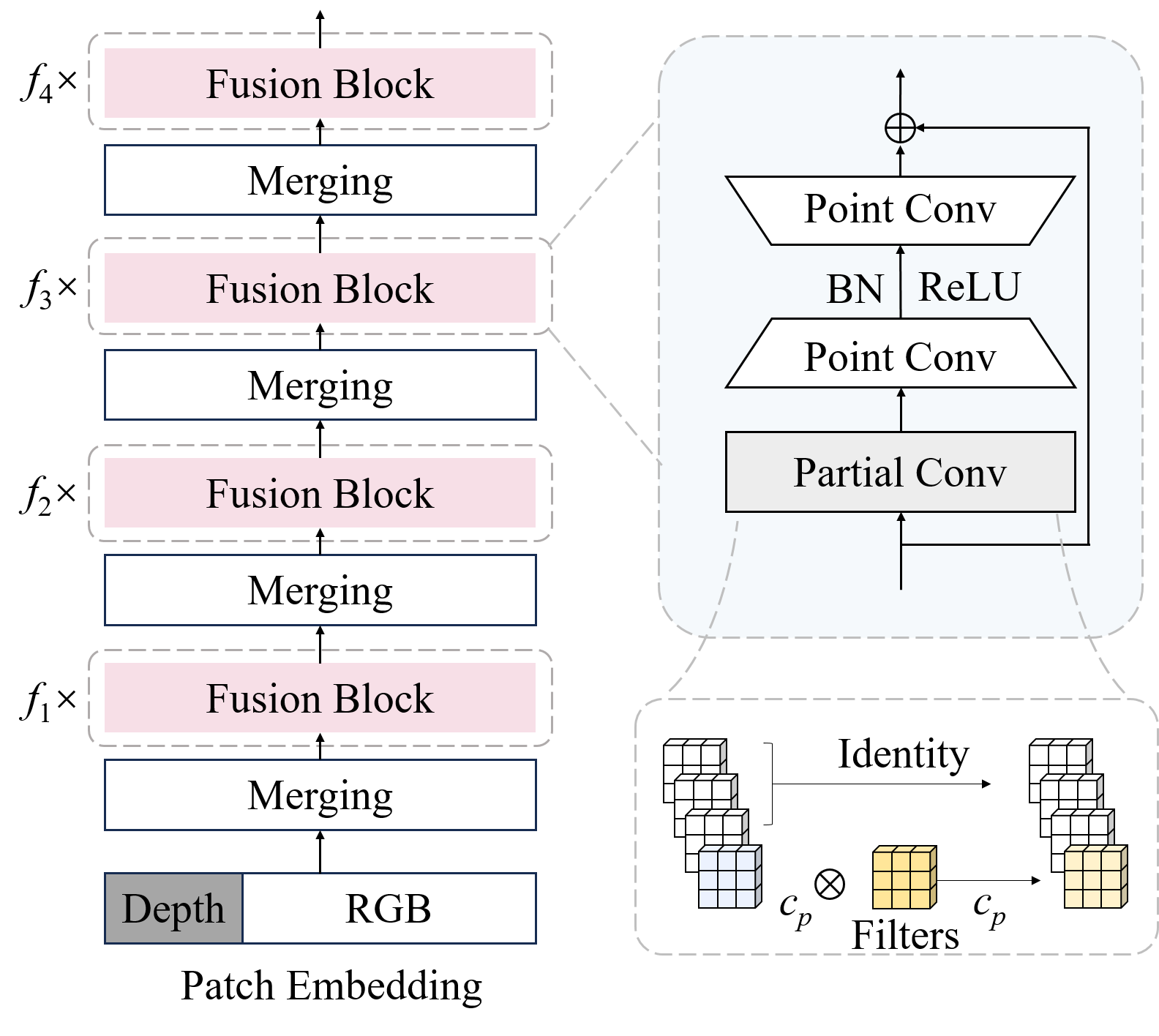}
	\caption{Fusion encoder. This encoder adopts merging layers for channel expansion and downsampling. Given the similarity between channel features, the fusion blocks extract redundant channel features to optimize computational efficiency.}
	\label{encoder}
\end{figure}

We employ an improved fusion encoder illustrated in Fig. \ref{encoder}. This encoder has 4 stages and processes input with RGBD channels. Each stage utilizes a $4\times4$ convolution with a normalization layer for channel expansion and downsampling, followed by several lightweight fusion blocks for feature extraction. As the image size decreases, memory access reduces, allowing us to increase the number of fusion blocks in the latter stages. Specifically, stages 1 through 4 contain 3, 4, 18, and 3 fusion blocks, respectively. In the fusion block, given an input feature $I\in\mathbb{R}^{C\times H\times W}$, where $C$ is the number of channels. Due to the high similarity of features across different channels, in order to fully exploit this redundancy, we select $1/4$ of the channels from $I$ as $I_{1}{\in}\mathbb{R}^{C^{\prime}\times H\times W}$, and denote the remaining channels as $I_{2}$. Feature extraction is then performed on the partial channel feature $I_{1}$, and then concatenates it with the remaining channels $I_{2}$:
\begin{equation}
F=\mathrm{Cat}((\mathrm{Conv2d}(I_1),I_2),\mathrm{dim}=1).
\end{equation}
The FLOPs of the partial convolution:
\begin{equation}
\mathrm{FLOPs}=H{\times}W{\times}k^2{\times}C^{\prime2},
\end{equation}
where $k$ is the convolution kernel size. Since $C^{\prime}=1/4~C$, partial convolution's FLOPs are reduced to $1/16$ of the regular convolution operation. Further, two pointwise convolutions are used to extract channel relationships. To enhance the representation of detailed scene features, the number of channels is initially expanded through pointwise convolution and then restored to its original count. Lastly, a residual connection is utilized to superimpose the original features. Due to the reduced memory access requirement, the encoder demonstrates a significant improvement in inference speed.

\subsection{Cross-dimensional feature guidance}\label{decoder}
\subsubsection{Normalized focus channel layer}

\begin{figure}[!t]
	\centering
	\includegraphics[width=3.2in]{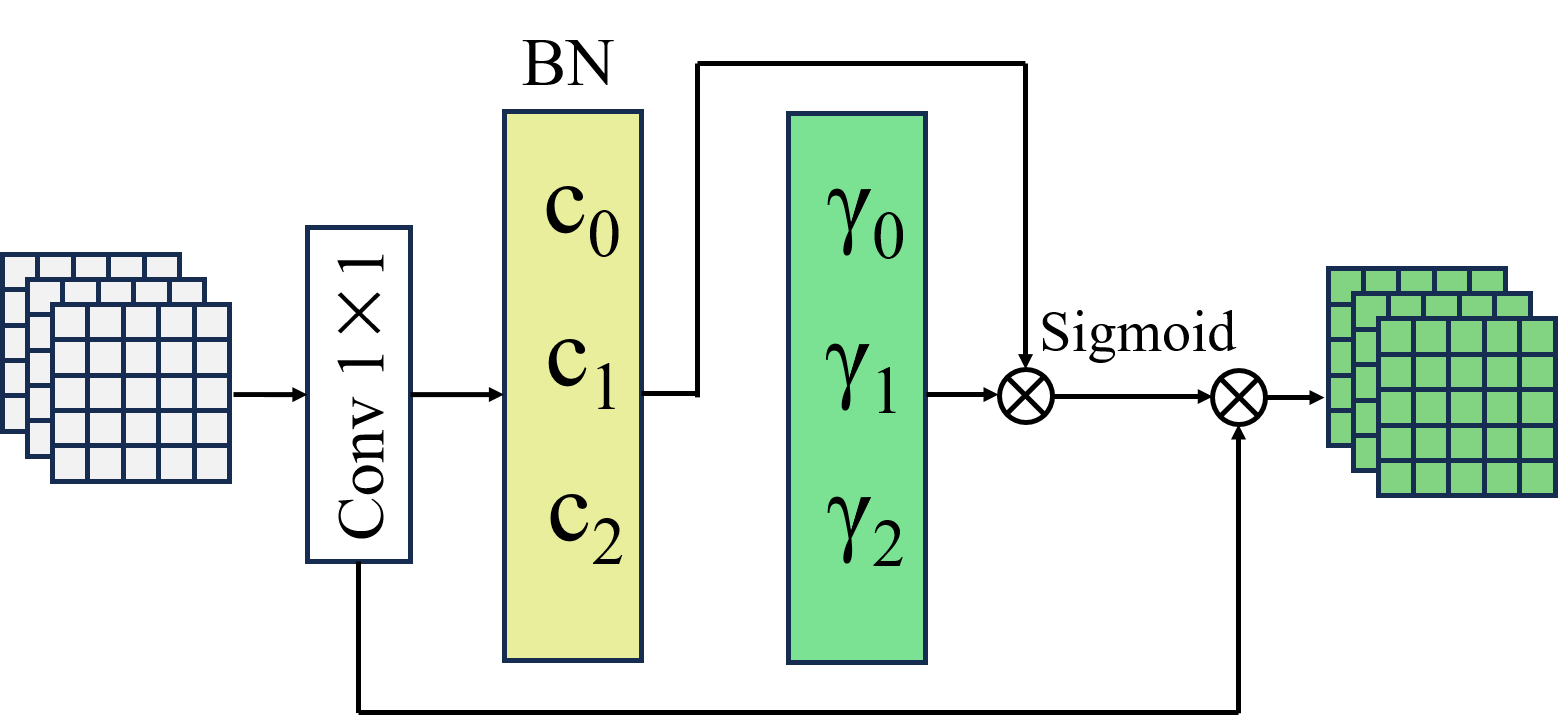}
	\caption{Normalized focus channel layer. This layer learns the variance adjustment parameters via batch normalization. It then computes channel weights that quantify the importance of each channel. The input features are rearranged, weighted according to these channel weights, and subsequently processed through an activation function.}
	\label{nfcl}
\end{figure}

In the semantic branch, the MLP-based decoder features a simpler structure compared to the traditional decoder with numerous convolutional and upsampling layers. This lightweight design enhances the model's efficiency during training and inference while reducing the risk of overfitting. However, the drawback is that erroneous information from the shallow features of the encoder can mislead the MLP, impacting the representation of local details \cite{shallow}. To address this, we design a normalized focus channel layer (NFCL) to enhance shallow information in the channel dimension. As depicted in Fig. \ref{nfcl}, the channel operation begins with feature extraction using a 1 $\times$ 1 convolution to obtain feature $X\in\mathbb{R}^{B\times C\times H\times W}$, where $B$ is the batch size, $C$ is the number of channels, and $H$, $W$ are the height and width of the feature map. Then, we perform the batch normalization operation:
\begin{equation}
\mathrm{BN}(X)=\gamma(\frac{X-\mu_B}{\sqrt{\sigma_B^2+\epsilon}})+\beta,
\end{equation}
where $\mu_B$ and $\sigma_B^2$ are the mean and variance over the batch. $\epsilon$ is a constant used to avoid division by zero and numerical instability. $\gamma$ and $\beta$ are learnable parameters used to scale and shift the normalized output. The scaling factor $\gamma$ corresponds to the variance in BN. A larger $\gamma$ indicates greater variation in the channel, suggesting it contains more important information \cite{nam}. Subsequently, we obtain the channel weight vector $\mathrm{W}_i$ by taking the absolute value of the scaling factor $\gamma$ and normalizing it:
\begin{equation}
\mathrm{W}_i=\frac{|\gamma_i|}{\sum_{j=1}^C|\gamma_j|},
\end{equation}
where $i$ denotes the channel index. The input feature $X\in\mathbb{R}^{B\times C\times H\times W}$ is then rearranged to $\widehat{X}\in\mathbb{R}^{B\times H\times W\times C}$ to align with the weight $\mathrm{W}$. Subsequently, the channel features at each pixel position are multiplied by their corresponding weights: $\tilde{X}=\mathrm{W}\cdot\hat{X}$. Finally, we apply a sigmoid activation function to the weighted features, then perform a pixel-wise multiplication with the original inputs to generate the enhanced features.

\subsubsection{Context feature interaction layer}
\begin{figure}[!t]
	\centering
	\includegraphics[width=3.6in]{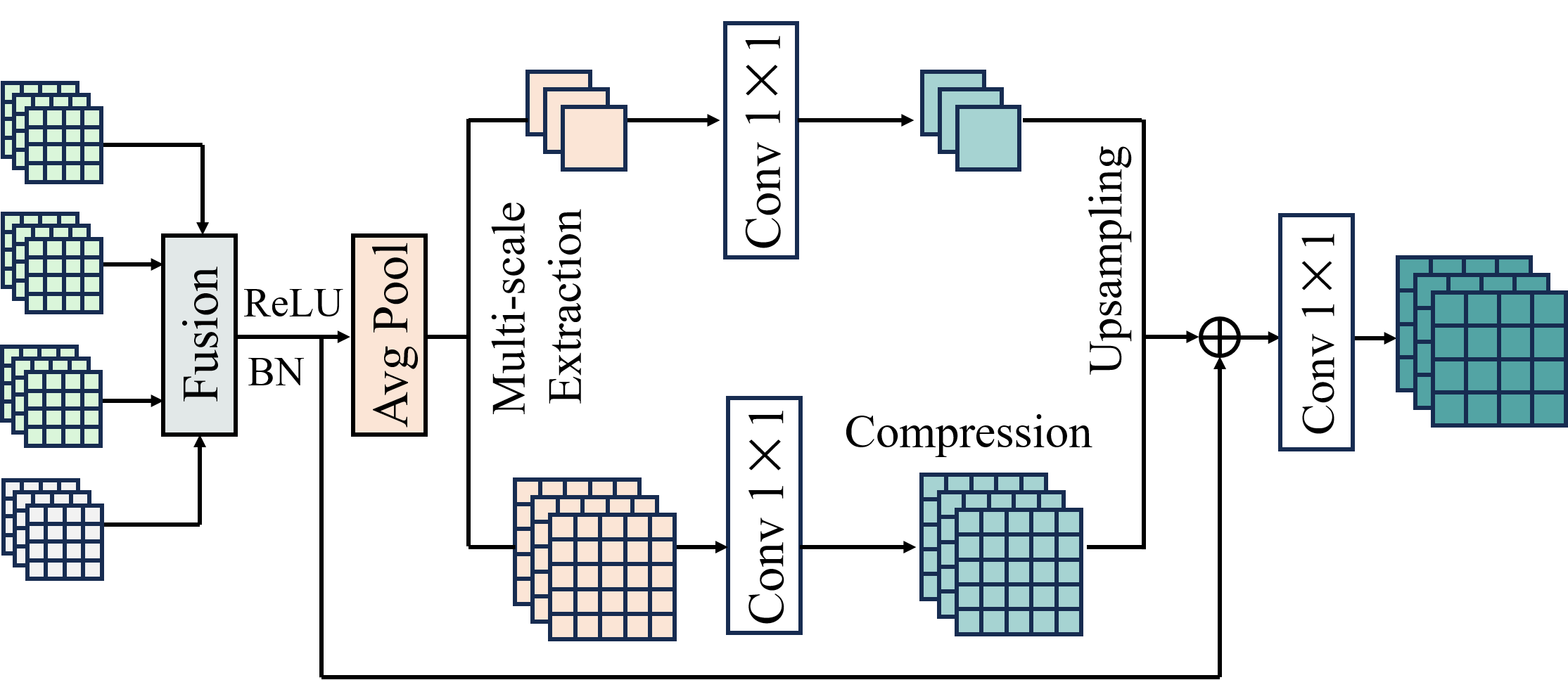}
	\caption{Context feature interaction layer. This layer addresses the limitation of the semantic decoder in integrating local and global information by employing a multi-scale pooling operation and efficient channel compression.}
	\label{cfsl}
\end{figure}

While the MLP-based semantic decoder excels at nonlinear feature mapping, it primarily focuses on the integration of global features \cite{segformer}. We design a context feature interaction layer (CFIL) to compensate for the deficiencies of the semantic decoder in fusing local and global information. The CFIL captures context information through multi-scale pooling operations. This approach allows the model to integrate features from different resolutions, improving its ability to discern and segment intricate structures and boundaries. In Fig. \ref{cfsl}, the CFIL concatenates and fuses the features from the NFCL and the convolutional layer to obtain feature $X\in\mathbb{R}^{B\times C\times H\times W}$. It then applies adaptive average pooling with predefined scales (1 $\times$ 1 and 5 $\times$ 5) to extract features at different scales. Each position $(i,j)$ of the output feature is computed as follows:
\begin{equation}
Y(i,j)=\frac{1}{k_h\cdot k_w}\sum_{m=0}^{k_h-1}\sum_{n=0}^{k_w-1}X(s_h\cdot i+m,s_w\cdot j+n).
\end{equation}

\begin{figure*}[!t]
	\centering
	\begin{subfigure}[!t]{0.14\textwidth}
		\includegraphics[width=0.95\linewidth]{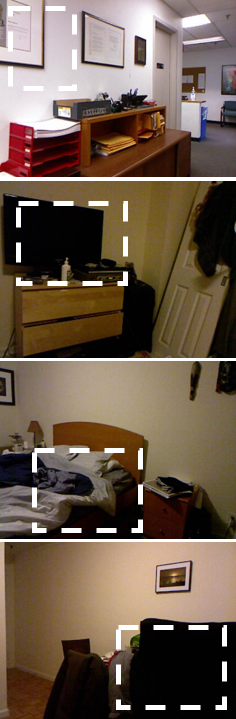}
		\caption{Inputs}
	\end{subfigure}
	\hspace{-0.7em}
	\begin{subfigure}[!t]{0.14\textwidth} 
		\includegraphics[width=0.95\linewidth]{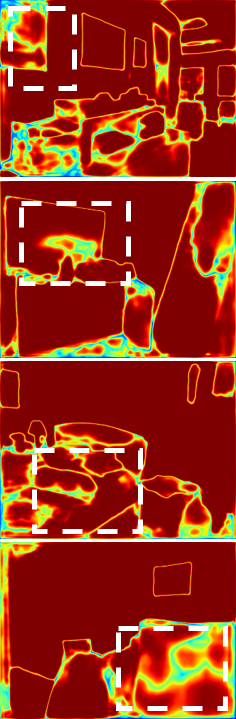}
		\caption{EMSAFormer}
	\end{subfigure}
	\hspace{-0.7em}
	\begin{subfigure}[!t]{0.14\textwidth} 
		\includegraphics[width=0.95\linewidth]{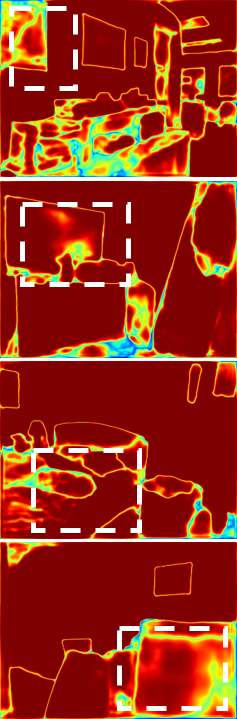}
		\caption{CFIL}
	\end{subfigure}
	\hspace{-0.7em}
	\begin{subfigure}[!t]{0.14\textwidth}
		\includegraphics[width=0.95\linewidth]{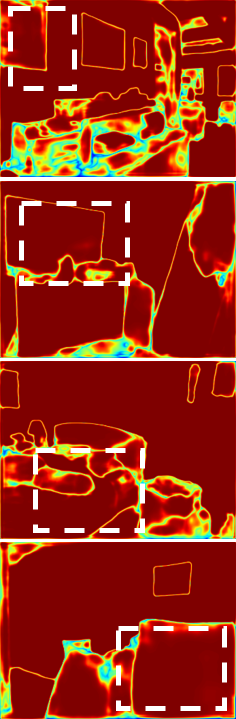}
		\caption{CFIL+NFCL}
	\end{subfigure}
	\caption{Semantic segmentation heat maps. Combining the feature guidance of CFIL and NFCL enables better handling of objects with similar colors and low-light scenes.}
	\label{semantic}
\end{figure*}

The $k_h$ and $k_w$ denote the predefined scales of the pooling kernel in the height and width dimensions. The strides are set as $s_h=k_h$ and $s_w=k_w$.
After obtaining the pooled features $Y_1\in\mathbb{R}^{B\times C\times1\times1}$ and $Y_2\in\mathbb{R}^{B\times C\times5\times5}$ through the above operation, a convolutional layer is applied to perform channel compression to reduce the dimensionality from $C$ to $C/2$. Bilinear interpolation upsampling is employed to unify the features from different scales into a consistent spatial resolution $(H \times W)$. Then, the upsampled features are concatenated with the original input $X$ to obtain the combined features $\widehat{Y}\in\mathbb{R}^{B\times(C+C/2+C/2)\times H\times W}$. Finally, $\widehat{Y}$ is adjusted to the original channel dimension $C$ through a convolutional layer, producing the refined output features. This design not only effectively captures and integrates multi-scale context information but also improves computational efficiency and feature expressiveness through a flexible upsampling approach and channel compression. To showcase the effectiveness of this method, we visualize the heat maps of the semantic segmentation branch in Fig. \ref{semantic}. By leveraging the guidance of CFIL and NFCL, the model more effectively handles objects with similar colors and in low-light environments. This is demonstrated by its improved segmentation of paintings with colors similar to the wall and the black television in dim lighting.

\subsubsection{Non-bottleneck 1D instance decoder}

\begin{figure}[!t]
	\centering
	\includegraphics[width=3.6in]{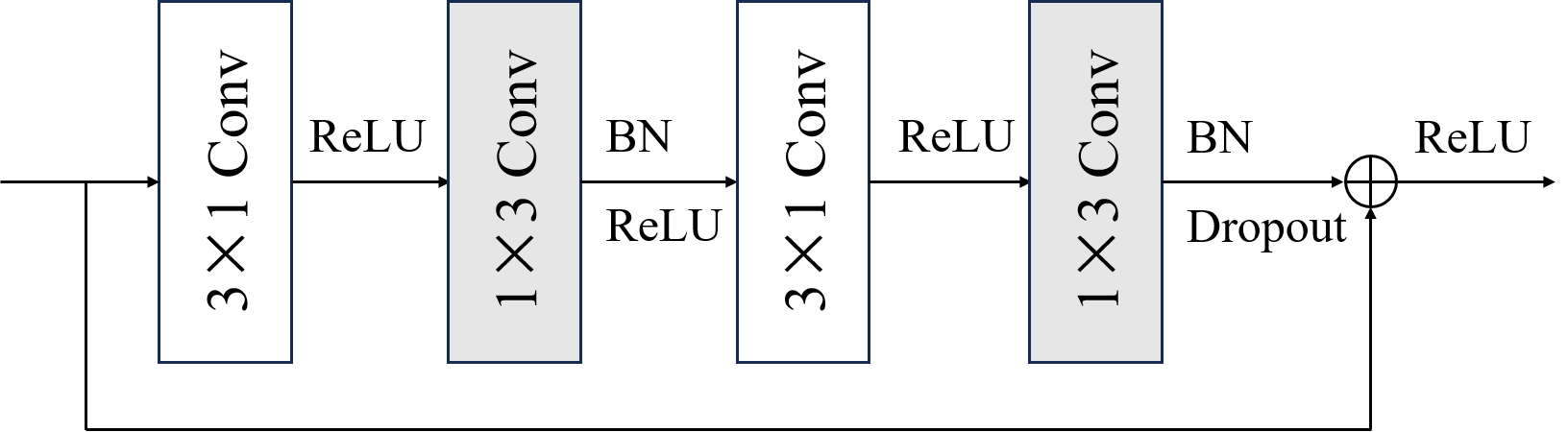}
	\caption{Non-bottleneck 1D module. This module decomposes 2D convolution into 1D convolutions combined with a non-linear activation function. This decomposition reduces model redundancy and enhances the non-linear decision-making capability.}
	\label{nonbottle}
\end{figure}
In the instance segmentation task, the complexity of simultaneously achieving instance segmentation and orientation estimation requires us to adopt a deeper decoder to obtain high-quality scene representation. To address this, we utilize a three-layer structured instance decoder. Each layer performs feature extraction through $3\times3$ convolution and combines three non-bottleneck 1D modules with an upsampling operation for refined feature extraction and resolution recovery. Inspired by Romera et al. \cite{nonbottle1d}, who proposed decomposing 2D convolution into two 1D convolutions with a non-linear activation function between them, we enhance learning capability while reducing the number of parameters. For a 2D convolution with the weight $W_{2D}\in\mathbb{R}^{C\times k\times k\times F}$, $C$ and $F$ represent the input and output channels, respectively, while $k$ denotes the kernel size. We decompose it into two 1D convolutions $W_{1D}\in\mathbb{R}^{C\times k\times F}$, reducing the parameter count to $2\times(C\times k\times F)$. Specifically, when the kernel size is 3, this approach reduces the parameters by 30$\%$ compared to a 2D convolution. As illustrated in Fig. \ref{nonbottle}, the non-bottleneck 1D module replaces each $3\times3$ convolution with a combination of $3\times1$ and $1\times3$ convolutions:

\begin{equation}
\mathrm{Conv}2\mathrm{D}_{(3,3)}=\mathrm{Conv}1\mathrm{D}_{(1,3)}
\begin{pmatrix}
\text{Activation}\left(\mathrm{Conv}1\mathrm{D}_{(3,1)}(X,W_1,b_1)\right),W_2
\end{pmatrix}.
\end{equation}

The $W$ and $b$ denote the weights and biases of the convolutional layers, respectively, and the activation function is ReLU.
This combination of 1D convolutions minimizes model redundancy, resulting in fewer parameters and increased computational efficiency.

\subsection{Multi-task adaptive learning}
Multi-task learning seeks to optimize multiple related tasks concurrently by exploiting shared representations and cross-task knowledge transfer, thereby enhancing the model's overall performance. Nevertheless, inherent variations in task complexity often result in imbalanced learning, where certain tasks dominate the training process while others receive insufficient attention \cite{kendall2018multi}. This challenge is further compounded by significant disparities in sample size and data quality. Moreover, the inter-task relationships exhibit dynamic characteristics that evolve throughout the training process as the model's learning state progresses. Conventional approaches employing static loss weights frequently fail to accommodate these temporal variations, thereby constraining the model's capacity to leverage evolving task correlations. To address this challenge, we propose a multi-task adaptive learning method that dynamically adjusts loss weights based on each task's historical performance. This methodology facilitates the balancing of cross-task influences, promoting more stable optimization dynamics and enhancing the model's responsiveness throughout the learning process.

\begin{algorithm}[!t]  
\caption{Multi-task adaptive learning} 
\setlength{\baselineskip}{1.3\baselineskip}
\textbf{Tasks:} Semantic segmentation, Instance center, Instance offset, Orientation estimation, Scene classification;\\
\textbf{Losses}: $L_{se}$,~~~$L_{ce}$,~~~$L_{of}$,~~~$L_{or}$,~~~$L_{sc}$;\\
Initialize the weights $\overline{W}$;\\
Sum the losses;\\
\ForEach{$k \in \mathrm{Tasks}$}{
Calculate the relative loss of Task $k$;\\
Update the relative loss history of Task $k$;\\
Get the historical average relative loss $\mathrm{Avg}RL_k$;\\
Update weight $W_k$ through the $\mathrm{Avg}RL_k$ and adjustment factor $\alpha$.}
Weight the losses with $W$.\\
\textbf{Output}: The sum of the weighted losses.
\label{algorithm}
\end{algorithm}


The specific flow is outlined in Algorithm \ref{algorithm}. At the end of each training batch, we evaluate the performance of the model across multiple tasks, including semantic segmentation, instance center prediction, instance offset prediction, orientation estimation, and scene classification, by calculating the respective loss values. To balance the impact of each task on the overall learning process, we first compute the relative loss $RL_k$ for each task $k$ with respect to the total loss, defined as:
\begin{equation}
RL_k=\frac{L_k}{\sum_tL_t},
\end{equation}
where $L_k$ denotes the loss for task $k$ and $\sum_tL_t$ is the sum of losses across all tasks. For each task $k$, we obtain an average of its historical relative losses to capture the task’s performance trend:
\begin{equation}
\mathrm{Avg}RL_k=\frac{\sum_iRL_k^{(i)}}{n_k}.
\end{equation}
The $n_k$ represents the number of historical relative losses recorded for task $k$. Based on the average relative loss, we adjust the loss weight for each task using an adjustment factor $\alpha$, with the updated weight $W_k$ computed as follows:
\begin{equation}
W_k=\max\left(\overline{W}_k\times(\mathrm{Avg}RL_k)^\alpha,W_{min}\right).
\end{equation}
The $\overline{W}_k$ is the initial weight for task $k$, $\alpha$ is set to 0.01 to control the sensitivity of weight adjustment, and $W_{min}$ is a minimum threshold weight set to 0.1 to prevent a task from being neglected due to excessively low weight resulting from outlier losses. Once the weights are updated, we use them to recalculate the weighted losses for each task. The final total loss is then obtained by summing these weighted losses:
\begin{equation}
L=\sum_kW_k\times L_k.
\end{equation}
This dynamic adjustment mechanism ensures that the model balances the contributions of each task more effectively, adapting to variations in task performance. The loss functions for each task are detailed as follows:

\textbf{Semantic segmentation:} This task is typically framed as a pixel-wise classification problem, where the goal is to assign each pixel in an image to a predefined semantic category. The adopted loss function for this task is the cross-entropy loss \cite{fcn}, which quantifies the discrepancy between the predicted probability distribution and the ground truth labels:
\begin{equation}
L_{se}=-\sum_{i=1}^N\sum_{c=1}^Ky_{ic}\mathrm{log}\left(h_\theta(x_i)_c\right).
\end{equation}
Here, $N$ is the number of samples, $K$ is the number of classes, $x_i$ represents the $i-th$ sample, $h_\theta(x_i)_c$ is the predicted probability of sample $x_i$ belonging to class $c$, and $y_{ic}$ denotes the ground truth label for sample $x_i$. 

\textbf{Instance segmentation:} This task uses foreground masks from semantic segmentation and requires the computation of instance centers and pixel offsets relative to the instance centers. The instance center loss $L_{ce}$ adopts mean squared error (MSE) \cite{centermask}:
\begin{equation}
L_{ce}=\frac{1}{N}\sum_{i=1}^{N}(y_{i}-\hat{y}_{i})^{2},
\end{equation}
where $\hat{y}_i$ is the predicted center for the $i-th$ sample and $y_{i}$ is the corresponding ground truth. The MSE penalizes large errors in instance center prediction, encouraging the model to refine its accuracy. The instance offset loss $L_{of}$ uses mean absolute error (MAE) \cite{fasterrcnn}:
\begin{equation}
L_{of}=\frac{1}{N}\sum_{i=1}^{N}|y_{i}-\hat{y}_{i}|.
\end{equation}
The MAE provides more stable gradients, which helps the model maintain consistent performance, especially in complex scenes with extreme values.

\textbf{Orientation estimation:} Orientation estimation calculates the angle of objects in the scene relative to the axis perpendicular to the ground. Considering that angles are continuous and periodic, a probability distribution model \cite{orientation} based on the cosine and sine vector $f=(\cos\varphi,\sin\varphi)$ is employed to represent the angle $\varphi$. 
This model is a continuous probability distribution model on the circle, which can avoid discontinuities in the loss. The orientation loss $L_{or}$ is defined as:
\begin{equation}
L_{or}=1-e^{\kappa(f\cdot t-1)},
\end{equation}
where $t$ is the target value, and $\kappa$ serves as a hyperparameter that controls the penalty for large deviations between predicted and target values.

\textbf{Scene classification:} Scene classification aims to predict the scene category of the entire input image. For this task, we employ the cross-entropy loss $L_{sc}$ specifically designed for supervised classification tasks \cite{places}. The loss is defined as:
\begin{equation}
L_{sc}=-\frac1N\sum_{i=1}^N\log\left(\frac{e^{z_{i,y_i}}}{\sum_{j=1}^Ce^{z_{i,j}}}\right),
\end{equation}
where $N$ is the number of samples, $C$ is the number of classes, $y_i$ is the true class label for the $i-th$ sample, and $z_{i,j}$ represents the model’s score for the $j-th$ class of the $i-th$ sample. 

For panoptic segmentation, which combines semantic and instance segmentation, no additional loss functions are computed during training. Instead, segmentation quality is assessed during the validation phase. By employing these loss functions and dynamically adjusting weights based on historical performance, the model achieves more flexible and effective multi-task learning.
\section{Experiment}
\subsection{Experimental detail}
\subsubsection{Parameter setting}
The experiments are conducted using the NYUv2 \cite{nyuv2}, SUN RGB-D \cite{sunrgbd}, and Cityscapes \cite{cityscapes} datasets. Training is performed on a single RTX 3090 Ti GPU. The software environment consists of PyTorch 2.0.1, Torchvision 0.15.2, CUDA 12.1, and Python 3.8.18. The model is optimized using the SGD optimizer, with a learning rate of 0.03, weight decay of $1\times10^{-4}$, and momentum of 0.9. For the NYUv2 dataset, the input resolution is set to $640\times480$, with a training and validation batch size of 8. The input resolution and training batch size for SUN RGB-D are identical to those of NYUv2, while the validation batch size is increased to 16. For the Cityscapes dataset, the input resolution is $1024\times512$, and both the training and validation batch sizes are set to 6. The results for EMSAFormer \cite{emsaformer} are obtained through experimental replication. The backbone of our model is based on FasterNet-M \cite{fasternet}. Comparative experiments include MetaFormer-M \cite{poolformer}, MPViT-Base \cite{mpvit}, Swin Transformer v2-T \cite{swinv2}, and ConvNeXt v2-Base \cite{convnextv2}.
\subsubsection{Dataset}
\textbf{NYUv2} \cite{nyuv2} dataset contains 795 training samples and 654 validation samples. It includes scene classification and 40 semantic categories. The ceiling, floor, and wall are considered background classes, while the remaining categories are treated as object classes. To avoid misguidance from small instances, the instances are restricted to cover at least 0.25$\%$ of the area. Due to the lack of orientation labels in the original NYUv2 dataset, we utilize orientation annotations provided by EMSANet \cite{emsanet}.

\textbf{SUN RGB-D} \cite{sunrgbd} dataset includes 5,285 training samples and 5,050 validation samples. This dataset provides scene classification and 37 semantic categories but lacks instance-level segmentation and orientation labels. The original SUN RGB-D dataset provides 3D bounding boxes with class labels and orientations. First, establish a mapping between class labels and 3D bounding boxes to obtain instance information. Then, match the instance information with the 3D semantic point clouds to extract instance segmentation masks and orientations.

\textbf{Cityscapes} \cite{cityscapes} dataset is a large-scale dataset designed for urban scene understanding. It contains 5,000 annotated images collected from 50 different cities, with 2,975 images for training, 500 for validation, and 1,525 for testing. Each image has a resolution of 2048 $\times$ 1024 pixels, providing fine-grained details for tasks such as semantic segmentation, instance segmentation, and object detection.
\subsubsection{Evaluation metrics}
To comprehensively assess performance, we use multiple evaluation metrics. For instance segmentation task, we employ panoptic quality (PQ) as the primary metric, which combines segmentation quality (SQ) and recognition quality (RQ) to measure the model’s ability to distinguish between different instances and categories. Additionally, for orientation estimation, we introduce the mean absolute angle error (MAAE), which calculates the average absolute error between predicted and true angles. In the semantic segmentation task, we use mean intersection over union (mIoU) to evaluate the degree of matching between predicted and ground-truth segmentation maps. For the scene classification task, we select balanced accuracy (bAcc) to ensure the model performs well even with imbalanced class distributions. In the panoptic segmentation task, we integrate mIoU, PQ, RQ, SQ, and MAAE to form a comprehensive and detailed evaluation system.

\subsection{Ablation study}
To validate the model's effectiveness, we conduct ablation experiments on the NYUv2 dataset. We compare the improved fusion encoder with existing encoders and then separately evaluate the CFIL, NFCL, non-bottleneck 1D layer, and multi-task adaptive loss. Finally, we perform an ablation study on the entire model.
\subsubsection{Fusion encoder}

\begin{table*}[!t]
\centering
\setlength{\tabcolsep}{10pt}
\renewcommand{\arraystretch}{1.1}
\caption{Encoder comparison results on the NYUv2 dataset. The symbols $\uparrow$ and $\downarrow$ indicate whether higher or lower values are better, respectively. The best metric is indicated in \textbf{bold}.}
\footnotesize
\begin{tabular}{lcc|c@{\hspace{2em}}c@{\hspace{2em}}c@{\hspace{2em}}c@{\hspace{2em}}c|cc}
\toprule
\multirow{2}{*}{Method} & \multicolumn{2}{c}{Instance} & \multicolumn{5}{c}{Panoptic} & Scene & Semantic \\
                        & PQ$\uparrow$                  & MAAE$\downarrow$               & mIoU$\uparrow$  & PQ$\uparrow$    & RQ$\uparrow$    & SQ$\uparrow$    & MAAE$\downarrow$ & bAcc$\uparrow$                & mIoU$\uparrow$                 \\ \midrule
Swin v2 \cite{swinv2}                & 58.49               & 21.09               & \textbf{50.51}  & \textbf{43.08}  & \textbf{52.22}  & \textbf{81.34}  & 20.46 & \textbf{77.11}                & \textbf{49.76}                 \\
ConvNeXt v2 \cite{convnextv2}            & 41.04               & 31.24               & 27.20  & 18.21  & 24.11  & 72.20   & 29.29 & 61.79                & 27.69                 \\
MPViT \cite{mpvit}                  & 57.77               & 21.18               & 47.09  & 39.43  & 47.78  & 81.13  & 19.82 & 75.16                & 47.44                 \\
MetaFormer \cite{mpvit}             & 53.31               & 23.69               & 44.75  & 34.01  & 42.96  & 77.86  & 22.10 & 74.51                & 43.27                 \\
Ours                    & \textbf{58.59}               & \textbf{18.67}               & 47.37  & 40.35  & 49.18  & 81.03  & \textbf{17.88} & 74.67                & 46.83                 \\ \bottomrule
\end{tabular}
\label{tab：backbone}%
\end{table*}%

\begin{figure*}[!t]
	\centering
	\begin{subfigure}[!t]{0.14\textwidth}
		\includegraphics[width=0.95\linewidth]{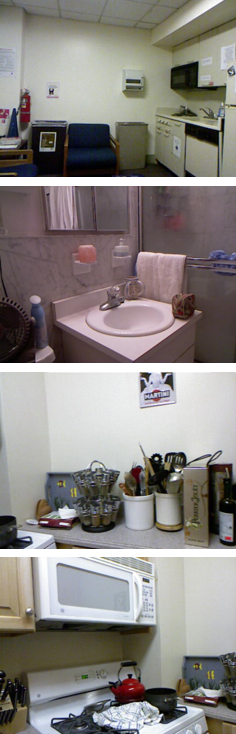}
		\caption{RGB}
	\end{subfigure}
	\hspace{-0.7em}
	\begin{subfigure}[!t]{0.14\textwidth} 
		\includegraphics[width=0.95\linewidth]{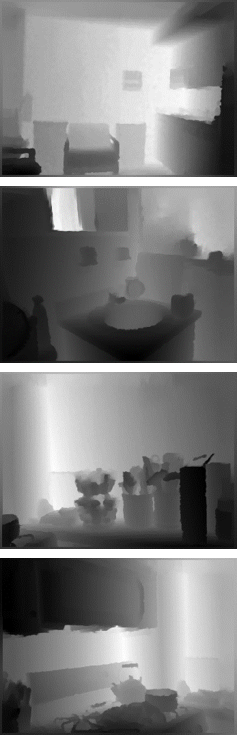}
		\caption{Depth}
	\end{subfigure}
	\hspace{-0.7em}
	\begin{subfigure}[!t]{0.14\textwidth} 
		\includegraphics[width=0.95\linewidth]{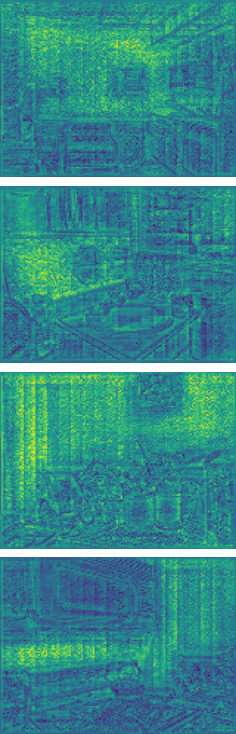}
		\caption{Swin v2}
	\end{subfigure}
	\hspace{-0.7em}
	\begin{subfigure}[!t]{0.14\textwidth}
		\includegraphics[width=0.95\linewidth]{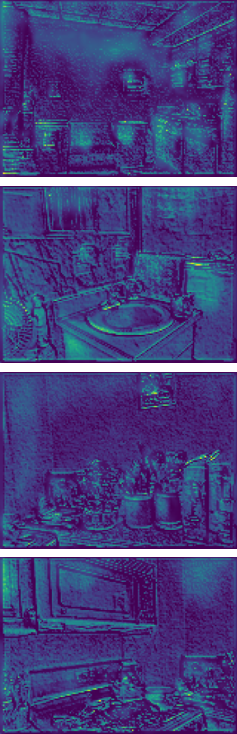}
		\caption{Ours}
	\end{subfigure}
	\caption{Feature visualization of backbones. We visualize the model inputs and the features extracted by Swin Transformer v2 and our method. Compared to the blurred features of Swin Transformer v2, our method more distinctly highlights objects in the scene, demonstrating better suitability for segmentation tasks.}
	\label{backbone}
\end{figure*}

The encoder plays a critical role in feature extraction from inputs. We compare the improved fusion encoder with both convolutional and transformer-based architectures in scene understanding tasks. As shown in Table \ref{tab：backbone}, our experiments reveal that the fully convolutional masked autoencoder ConvNeXt v2 falls short in dense prediction tasks, achieving an instance PQ of 41.04\%. Similarly, the MPViT, designed specifically for dense prediction, struggles with multitasking scenes. In contrast, the fusion encoder excels with an instance PQ of 58.59\%, outperforming other methods. While Swin Transformer v2 (Swin v2) shows strong performance, its frequent memory accesses lead to slower inference speed. Figure \ref{backbone} displays the feature maps of these encoders. Swin v2 broadly highlights the scene information, but the object boundaries are relatively blurred. In contrast, the improved fusion encoder more clearly emphasizes the object contours. This clarity of contours helps the model to more accurately segment target regions, thereby improving the performance of the segmentation task.

\subsubsection{Context feature interaction layer}

\begin{table*}[!t]
\centering
\setlength{\tabcolsep}{10pt}
\renewcommand{\arraystretch}{1.1}
\caption{Comparison of context modules. We compare the proposed context feature interaction layer with existing context modules. They are used in the multi-layer feature fusion stage of the semantic decoder.}
\footnotesize
\begin{tabular}{lcc|c@{\hspace{2em}}c@{\hspace{2em}}c@{\hspace{2em}}c@{\hspace{2em}}c|cc}
\toprule
\multirow{2}{*}{Method} & \multicolumn{2}{c}{Instance} & \multicolumn{5}{c}{Panoptic} & Scene & Semantic \\
                       & PQ$\uparrow$                  & MAAE$\downarrow$               & mIoU$\uparrow$  & PQ$\uparrow$    & RQ$\uparrow$    & SQ$\uparrow$    & MAAE$\downarrow$ & bAcc$\uparrow$                & mIoU$\uparrow$                  \\ \midrule
SPPELAN \cite{sppelanyolov9}                  & 58.79         & 19.18        & 47.55 & 41.09 & 49.60 & 81.66 & 18.18 & \textbf{77.39} & 46.51    \\
ASPP \cite{aspp}                    & 58.75         & 18.14        & 48.40 & 41.81 & 50.51 & 81.61 & 17.83 & 76.73 & 47.98    \\
SimSPPF \cite{simsppfyolov6}                 & 58.34         & 19.07        & 47.62 & 40.56 & 49.12 & 81.37 & 18.03 & 74.70 & 46.40    \\
RFB \cite{rfb}                     & 58.98         & \textbf{18.08}        & 48.46 & 42.15 & 50.83 & \textbf{81.88} & 17.75 & 76.51 & 47.14    \\
Ours                     & \textbf{59.25}         & 18.27        & \textbf{50.16} & \textbf{42.83} & \textbf{51.78} & 81.60 & \textbf{17.56} & 77.00 & \textbf{49.72}    \\ \bottomrule
\end{tabular}
\label{tab：context}%
\end{table*}%

\begin{table*}[!t]
\centering
\setlength{\tabcolsep}{10pt}
\renewcommand{\arraystretch}{1.1}
\caption{Comparison of the context feature interaction layer positions. P0: CFIL not used. P1: encoder. P2: instance decoder. P3: semantic decoder and instance decoder. P4: encoder and semantic decoder. Ours: semantic decoder.}
\footnotesize
\begin{tabular}{lcc|c@{\hspace{2em}}c@{\hspace{2em}}c@{\hspace{2em}}c@{\hspace{2em}}c|cc}
\toprule
\multirow{2}{*}{Method} & \multicolumn{2}{c}{Instance} & \multicolumn{5}{c}{Panoptic} & Scene & Semantic \\
                        & PQ$\uparrow$                  & MAAE$\downarrow$               & mIoU$\uparrow$  & PQ$\uparrow$    & RQ$\uparrow$    & SQ$\uparrow$    & MAAE$\downarrow$ & bAcc$\uparrow$                & mIoU$\uparrow$                  \\ \midrule
P0                               & 58.39         & 18.41        & 46.93 & 40.61 & 49.14 & 81.54 & 18.06 & 77.58 & 46.36    \\
P1                               & 59.37         & 19.34        & 48.39 & 41.81 & 51.02 & 80.78 & 18.58 & 76.23 & 47.72    \\
P2                               & 58.71         & \textbf{16.63}        & 48.17 & 41.44 & 50.21 & 81.55 & \textbf{15.63} & \textbf{77.89} & 47.12    \\
P3                           & 59.49         & 17.90        & 48.99 & 42.40 & 51.24 & 81.56 & 17.25 & 75.08 & 49.25    \\
P4                           & 58.88         & 19.78        & 47.62 & 40.52 & 49.05 & 81.60 & 18.41 & 75.21 & 47.13    \\
Ours                               & \textbf{59.25}         & 18.27        & \textbf{50.16} & \textbf{42.83} & \textbf{51.78} & \textbf{81.60} & 17.56 & 77.00 & \textbf{49.72}    \\ \bottomrule
\end{tabular}
\label{tab：contextlocation}%
\end{table*}%

The context feature interaction layer is designed to capture multi-scale context information, thereby enhancing the model's performance in complex scenes. Our study demonstrates the advantages of CFIL through comparisons with existing methods, as detailed in Table \ref{tab：context}. Specifically, the CFIL achieves optimal values across six key performance metrics, with semantic mIoU and instance PQ reaching 49.72\% and 59.25\%, respectively. This result indicates that CFIL not only improves the quality of semantic foreground masks but also indirectly enhances the accuracy of instance segmentation. Furthermore, we explore the impact of the position of feature interaction on various tasks in Table \ref{tab：contextlocation}. Due to the deeper layer in the instance branch, the additional feature interaction leads to limited improvements in panoptic quality. When CFIL is applied to the semantic branch, the model exhibits the most balanced performance, with a panoptic mIoU reaching 50.16\%.

\subsubsection{Normalized focus channel layer}
\begin{table*}[!t]
\centering
\setlength{\tabcolsep}{10pt}
\renewcommand{\arraystretch}{1.1}
\caption{Comparison of the normalized focus channel layers on the semantic decoder. L0: 1st, 2nd, 3rd, and 4th layers. L1: 2nd and 3rd layers. L2: 1st and 2nd layers. L3: 2nd, 3rd, and 4th layers. Ours: 1st, 2nd, and 3rd layers.}
\footnotesize
\begin{tabular}{lcc|c@{\hspace{2em}}c@{\hspace{2em}}c@{\hspace{2em}}c@{\hspace{2em}}c|cc}
\toprule
\multirow{2}{*}{Method} & \multicolumn{2}{c}{Instance} & \multicolumn{5}{c}{Panoptic} & Scene & Semantic \\
                        & PQ$\uparrow$                  & MAAE$\downarrow$               & mIoU$\uparrow$  & PQ$\uparrow$    & RQ$\uparrow$    & SQ$\uparrow$    & MAAE$\downarrow$ & bAcc$\uparrow$                & mIoU$\uparrow$                  \\ \midrule
L0                 & 59.15         & 19.03        & 49.37 & 42.00 & 50.61 & \textbf{81.97} & 17.74 & 76.48 & 48.93    \\
L1                   & 58.78         & 19.88        & 49.24 & 42.52 & 51.60  & 81.21 & 18.08 & 74.64 & 48.88    \\
L2                   & 59.37         & 18.42        & 50.01 & 42.74 & 51.71 & 81.5  & 17.78 & 76.42 & 49.53    \\
L3                  & 58.97         & \textbf{18.03}        & 48.41 & 41.65 & 50.32 & 81.76 & 17.41 & 76.35 & 48.45    \\
Ours                  & \textbf{59.90}         & 19.64        & \textbf{50.21} & \textbf{43.21} & \textbf{52.58} & 81.20  & 18.79 & \textbf{76.57} & \textbf{49.82}    \\ \bottomrule
\end{tabular}
\label{tab：nfcl}%
\end{table*}%

The primary objective behind the design of the normalized focus channel layer is to enhance the representation capability of information from the shallow encoder layers. We implement NFCL guidance in the skip connection layers of the semantic decoder, which consist of four layers in total. As shown in Table \ref{tab：nfcl}, applying NFCL to the 1st, 2nd, and 3rd layers yields better performance compared to applying it to the 2nd, 3rd, and 4th layers. This suggests that the 4th layer's encoder feature extraction is already effective and does not require additional guidance. Thus, placing NFCL in the 1st, 2nd, and 3rd layers optimizes the model's performance, achieving a semantic mIoU of 49.82\%.

\subsubsection{Non-bottleneck 1D decoder}
\begin{table*}[!t]
\centering
\setlength{\tabcolsep}{9pt}
\renewcommand{\arraystretch}{1.1}
\caption{Comparison of feature extraction modules. In the instance decoder layer, we compare the non-bottleneck 1D module with BasicBlock, Bottleneck, MobileBottleneck, and GhostBottleneck.}
\footnotesize
\begin{tabular}{lcc|c@{\hspace{2em}}c@{\hspace{2em}}c@{\hspace{2em}}c@{\hspace{2em}}c|cc}
\toprule
\multirow{2}{*}{Method} & \multicolumn{2}{c}{Instance} & \multicolumn{5}{c}{Panoptic} & Scene & Semantic \\
                        & PQ$\uparrow$                  & MAAE$\downarrow$               & mIoU$\uparrow$  & PQ$\uparrow$    & RQ$\uparrow$    & SQ$\uparrow$    & MAAE$\downarrow$ & bAcc$\uparrow$                & mIoU$\uparrow$                  \\ \midrule
BasicBlock \cite{resnet}                       & 58.51         & 18.40        & 49.33 & 42.29 & 51.17 & 81.45 & \textbf{17.40} & 76.87 & 49.17    \\
Bottleneck \cite{resnet}                       & 57.97         & 18.80        & \textbf{50.33} & 41.97 & 50.93 & 81.23 & 18.20 & 76.41 & \textbf{50.26}    \\
MobileBottleneck \cite{mobilenet}                 & 58.24         & 20.45        & 48.46 & 40.85 & 49.45 & 81.38 & 19.54 & 75.03 & 48.66    \\
GhostBottleneck \cite{ghostnet}                  & 58.14         & 20.29        & 48.93 & 41.82 & 51.02 & 80.74 & 19.72 & 75.86 & 48.75    \\
Ours                              & \textbf{59.25}         & \textbf{18.27}        & 50.16 & \textbf{42.83} & \textbf{51.78} & \textbf{81.60} & 17.56 & \textbf{77.00} & 49.72    \\ \bottomrule
\end{tabular}
\label{tab：nonbottle}%
\end{table*}%

In the instance decoder, we evaluate the effectiveness of the non-bottleneck 1D module by comparing it with several commonly used feature extraction modules. In Table \ref{tab：nonbottle}, GhostBottleneck \cite{ghostnet} employs the Ghost module to reduce computational costs, but this approach cannot fully utilize the correlations between features. MobileBottleneck \cite{mobilenet} relies heavily on depthwise separable convolution, leading to inadequate fusion of features across different layers. This limitation is particularly pronounced in dense prediction tasks. While the Bottleneck \cite{resnet} module minimizes parameters and computational complexity through dimensionality reduction and expansion, this process can result in information loss. In contrast, the non-bottleneck 1D module uses 1D convolutions directly for feature extraction, which preserves the original feature dimensions. As a result, it achieves superior performance across six metrics, with an instance PQ of 59.25\%.
\subsubsection{Multi-task adaptive loss}

\begin{figure*}[!t]
	\centering
	\begin{subfigure}[!t]{0.6\textwidth}
		\includegraphics[width=0.95\linewidth]{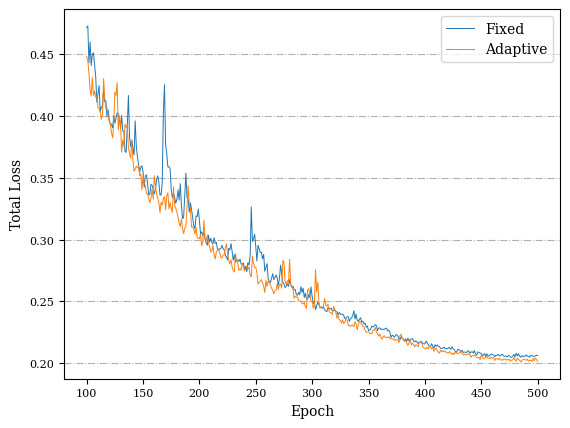}
		\caption{Loss comparison}
        \label{Loss}
	\end{subfigure}
	\begin{subfigure}[!t]{0.62\textwidth} 
		\includegraphics[width=0.95\linewidth]{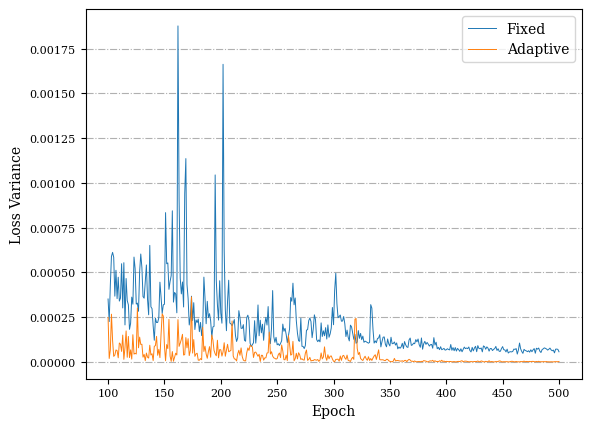}
		\caption{Loss variance}
        \label{variance}
	\end{subfigure}
	\caption{Comparison plot of multi-task adaptive loss. Figure a and Figure b represent the loss comparison and loss variance, respectively. The blue line represents the original fixed-weight loss, while the orange line represents the multi-task adaptive loss. To make the comparison clearer, we record the loss values and variances from epoch 100 onward. Our method makes the training process more convergent and stable.}
\end{figure*}

Multi-task adaptive loss dynamically adjusts the focus of the model's learning. The adjustment factor determines the extent of this adjustment. To select an appropriate adjustment magnitude, we conduct experiments with different adjustment factors. As shown in Table \ref{tab：loss}, decreasing the adjustment magnitude progressively improves the model’s learning effectiveness, demonstrating that multi-task learning benefits from fine-tuning. When the adjustment factor is 0.01, the model achieves the most balanced performance, with a semantic mIoU of 47.72\% and a panoptic PQ of 41.81\%. Additionally, we record the values of fixed-weight loss versus multi-task adaptive loss during training in Fig. \ref{Loss}. The fixed-weight loss exhibits significant fluctuations during training. In contrast, multi-task adaptive loss converges more quickly and smoothly. Meanwhile, we conducted five experiments for each of the two loss functions and calculated the variance for each epoch across the five experiments. As shown in Fig. \ref{variance}, the multi-task adaptive loss demonstrates better convergence, indicating smaller differences across experiments and greater stability.

\begin{table*}[!t]
\centering
\setlength{\tabcolsep}{10pt}
\renewcommand{\arraystretch}{1.1}
\caption{Comparison of adjustment factors for multi-task adaptive loss. "Fixed" represents the original fixed weight. The numbers denote the adjustment factors, with our method using an adjustment factor of 0.01.}
\footnotesize
\begin{tabular}{lcc|c@{\hspace{2em}}c@{\hspace{2em}}c@{\hspace{2em}}c@{\hspace{2em}}c|cc}
\toprule
\multirow{2}{*}{Method} & \multicolumn{2}{c}{Instance} & \multicolumn{5}{c}{Panoptic} & Scene & Semantic \\
                        & PQ$\uparrow$                  & MAAE$\downarrow$               & mIoU$\uparrow$  & PQ$\uparrow$    & RQ$\uparrow$    & SQ$\uparrow$    & MAAE$\downarrow$ & bAcc$\uparrow$                & mIoU$\uparrow$                  \\ \midrule
Fixed                 & 58.59         & \textbf{18.67}        & 47.37 & 40.35 & 49.18 & 81.03 & \textbf{17.88} & 74.67 & 46.83    \\
0.1                   & 57.66         & 19.90        & 47.17 & 40.05 & 49.05 & 80.63 & 19.15 & 72.99 & 47.21    \\
0.05                  & 58.86         & 19.06        & 47.72 & 40.07 & 48.74 & 80.96 & 18.33 & 73.59 & 46.74    \\
0.02                  & \textbf{59.40}         & 19.02        & 48.25 & 41.56 & 50.34 & \textbf{81.44} & 18.21 & 74.59 & 47.63    \\
0.01                  & 59.37         & 19.34        & \textbf{48.39} & \textbf{41.81} & \textbf{51.02} & 80.78 & 18.58 & \textbf{76.23} & \textbf{47.72}    \\ \bottomrule
\end{tabular}
\label{tab：loss}%
\end{table*}%

\subsubsection{Framework}

\begin{table*}[!t]
\centering
\setlength{\tabcolsep}{8pt}
\renewcommand{\arraystretch}{1.1}
\caption{
Comparison results of the ablation study on the overall framework. Fusion: improved fusion encoder. Adaptive: multi-task adaptive loss. CFIL: context feature interaction layer. NFCL: normalized focus channel layer.}
\footnotesize
\begin{tabular}{c@{\hspace{0.5em}}c@{\hspace{0.5em}}c@{\hspace{0.5em}}c@{\hspace{0.5em}}cc|c@{\hspace{2em}}c@{\hspace{2em}}c@{\hspace{2em}}c@{\hspace{2em}}c|cc}
\toprule
\multirow{2}{*}{Fusion} &\multirow{2}{*}{Adaptive} &\multirow{2}{*}{CFIL} &\multirow{2}{*}{NFCL}  & \multicolumn{2}{c}{Instance} & \multicolumn{5}{c}{Panoptic} & Scene & Semantic \\
                      & & &    & PQ$\uparrow$                  & MAAE$\downarrow$               & mIoU$\uparrow$  & PQ$\uparrow$    & RQ$\uparrow$    & SQ$\uparrow$    & MAAE$\downarrow$ & bAcc$\uparrow$                & mIoU$\uparrow$                  \\ \midrule
  & & &                             & 58.49         & 21.09        & \textbf{50.51} & 43.08 & 52.22 & 81.34 & 20.46 & \textbf{77.11} & 49.76    \\
 \checkmark   & &  &                            & 58.59         & 18.67       & 47.37 & 40.35 & 49.18 & 81.03 & 17.88 & 74.67 & 46.83    \\
 \checkmark   &  \checkmark &  &                               & 59.37        & 19.34        & 48.39 & 41.81 & 51.02 & 80.78 & 18.58 & 76.23 & 47.72    \\
 \checkmark   &  \checkmark & \checkmark &                           & 59.25         & \textbf{18.27}        & 50.16 & 42.83 & 51.78 & \textbf{81.60} & \textbf{17.56} & 77.00 & 49.72    \\
 \checkmark   &  \checkmark & \checkmark &   \checkmark            & \textbf{59.90}         & 19.64       & 50.21 & \textbf{43.21} & \textbf{52.58} & 81.20 & 18.79 & 76.57 & \textbf{49.82}    \\
\bottomrule
\end{tabular}
\label{tab：framework}%
\end{table*}%

\begin{figure*}[!t]
	\centering
	\begin{subfigure}[!t]{0.14\textwidth}
		\includegraphics[width=0.95\linewidth]{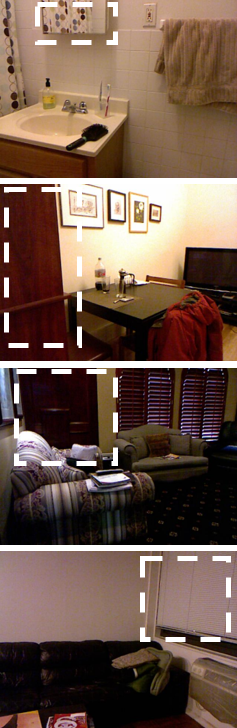}
		\caption{RGB}
	\end{subfigure}
	\hspace{-0.7em}
	\begin{subfigure}[!t]{0.14\textwidth} 
		\includegraphics[width=0.95\linewidth]{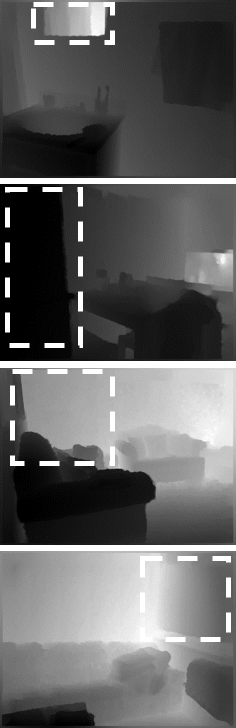}
		\caption{Depth}
	\end{subfigure}
	\hspace{-0.7em}
	\begin{subfigure}[!t]{0.14\textwidth} 
		\includegraphics[width=0.95\linewidth]{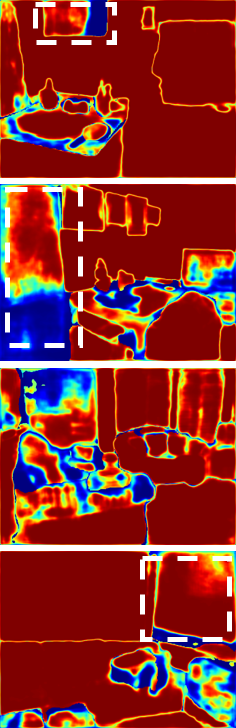}
		\caption{Baseline}
	\end{subfigure}
	\hspace{-0.7em}
	\begin{subfigure}[!t]{0.14\textwidth}
		\includegraphics[width=0.95\linewidth]{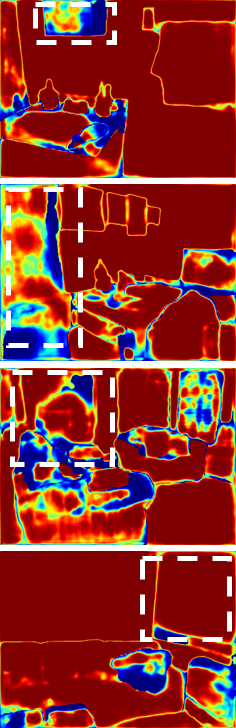}
		\caption{Fusion}
	\end{subfigure}
	\hspace{-0.7em}
	\begin{subfigure}[!t]{0.14\textwidth} 
		\includegraphics[width=0.95\linewidth]{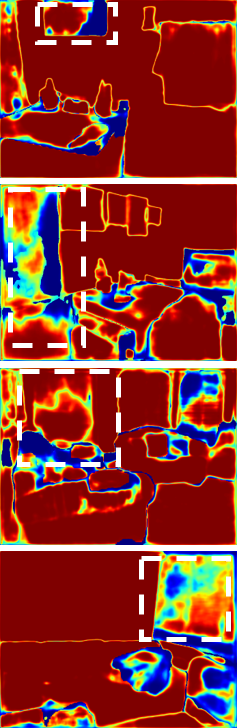}
		\caption{Adaptive}
	\end{subfigure}
	\hspace{-0.7em}
	\begin{subfigure}[!t]{0.14\textwidth} 
		\includegraphics[width=0.95\linewidth]{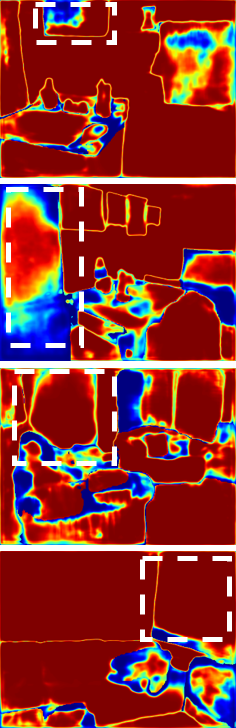}
		\caption{CFIL}
	\end{subfigure}
	\hspace{-0.7em}
	\begin{subfigure}[!t]{0.14\textwidth} 
		\includegraphics[width=0.95\linewidth]{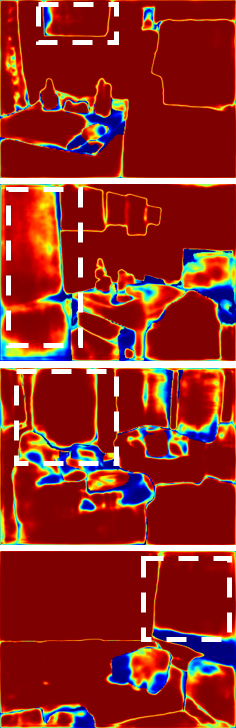}
		\caption{NFCL}
	\end{subfigure}
	\caption{Visualization of the ablation experiment. We present the panoptic heatmaps for the model with the improvements: fusion encoder, multi-task adaptive loss, context feature interaction layer, and normalized focus channel layer. Benefiting from the adaptive learning strategy and cross-dimensional feature guidance, the model achieves more accurate recognition and contour representation.}
	\label{Ablation}
\end{figure*}

To assess the overall framework's effectiveness, we conduct ablation experiments on each design component of the model, as shown in Table \ref{tab：framework}. Beginning with the baseline, we utilize the efficient fusion encoder, which improves feature fusion and enhances performance on the instance segmentation task. We then introduce multi-task adaptive loss, which dynamically adjusts the learning strategy for each task. This adjustment leads to improvements in six metrics, with instance PQ reaching 59.37\%. Next, we incorporate CFIL into the semantic decoder to capture multi-scale context information. This addition results in improvements across eight metrics, including a 2-point increase in semantic mIoU, reaching 49.72\%. Finally, we add NFCL to focus on local detail information, achieving gains in five metrics and a panoptic PQ of 43.21\%. Figure \ref{Ablation} visualizes the ablation experiment results with panoptic heatmaps. The fusion encoder effectively reduces focus variations within the same object range compared to the baseline. The introduction of the adaptive learning strategy enhances the model's ability to recognize furniture. Additionally, the cross-dimensional feature guidance from CFIL and NFCL strengthens the model's focus on stacked objects, leading to more accurate scene representation.

\subsection{Model complexity}
\begin{table}[width=0.98\linewidth,pos=!t]
\centering
\setlength{\tabcolsep}{12pt}
\renewcommand{\arraystretch}{1.1}
\caption{Comparison results of model complexity. Based on EMSAFormer, we compare the proposed model with models using different encoders. The comparison metrics include the number of parameters, FLOPs, FPS, and video memory usage (VRAM).}
\footnotesize
\begin{tabular}{lcccccc}
\toprule
Method     & Params & FLOPs  & FPS   &VRAM & Semantic mIoU$\uparrow$ & Instance PQ$\uparrow$\\ \midrule
EMSAFormer \cite{emsaformer}  & 72.08 M  & \textbf{50.66} G & 16.32 & \textbf{3188} MiB & 49.76  & 58.49       \\
ConvNeXt v2 \cite{convnextv2}& 111.84 M& 112.38 G& 14.42 & 4592 MiB& 27.69  & 41.04      \\
MetaFormer \cite{poolformer}& 76.14 M & 72.06 G & 12.01 & 3620 MiB& 43.27   & 53.31      \\
MPViT \cite{mpvit}& 92.76 M & 235.24 G & 9.94  & 5266 MiB& 47.44    & 57.77    \\
Ours       & \textbf{71.82} M  & 75.28 G & \textbf{20.33} & 3293 MiB& \textbf{49.82}    & \textbf{59.90}    \\ \bottomrule
\end{tabular}
\label{tab：complexity}%
\end{table}%
We compare the complexity of the proposed model with other models. As shown in Table \ref{tab：complexity}, our model achieves the lowest number of parameters at 71.82M, while maintaining low FLOPs and video memory usage. Benefiting from the utilization of redundant features, which reduces memory access frequency, our model achieves an FPS of 20.33, outperforming other methods. The transformer architecture has advantages over the convolutional architecture ConvNeXt v2 in terms of accuracy. However, as a convolutional architecture, the proposed model achieves the best precision, with a semantic mIoU of 49.82\% and an instance PQ of 59.90\%.

\subsection{Comparison with SOTA methods}
\subsubsection{NYUv2}

\begin{figure*}[!t]
	\centering
        \hspace{-0.7em}
	\begin{subfigure}[!t]{0.14\textwidth}
		\includegraphics[width=0.95\linewidth]{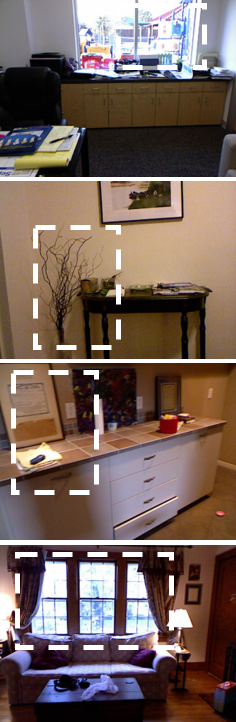}
		\caption{RGB}
	\end{subfigure}
	\hspace{-0.7em}
	\begin{subfigure}[!t]{0.14\textwidth} 
		\includegraphics[width=0.95\linewidth]{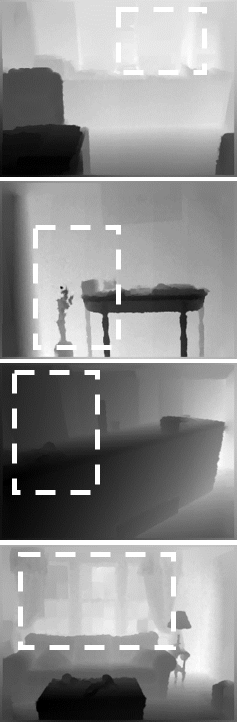}
		\caption{Depth}
	\end{subfigure}
	\hspace{-0.7em}
	\begin{subfigure}[!t]{0.14\textwidth} 
		\includegraphics[width=0.95\linewidth]{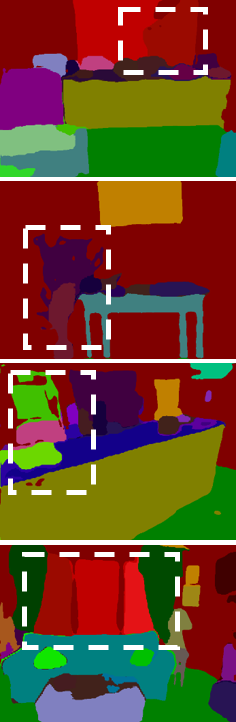}
		\caption{Swin v2}
	\end{subfigure}
	\hspace{-0.7em}
	\begin{subfigure}[!t]{0.14\textwidth}
		\includegraphics[width=0.95\linewidth]{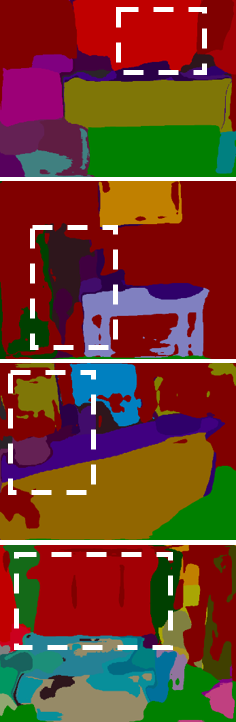}
		\caption{ConvNeXt v2}
	\end{subfigure}
	\hspace{-0.7em}
	\begin{subfigure}[!t]{0.14\textwidth} 
		\includegraphics[width=0.95\linewidth]{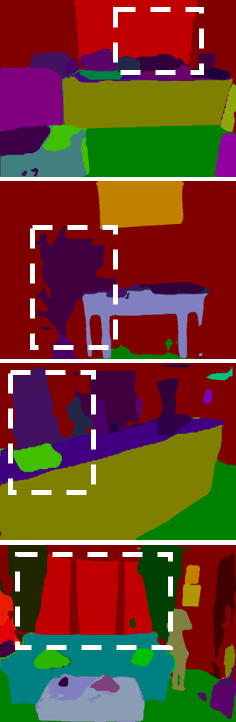}
		\caption{MPViT}
	\end{subfigure}
	\hspace{-0.7em}
	\begin{subfigure}[!t]{0.14\textwidth} 
		\includegraphics[width=0.95\linewidth]{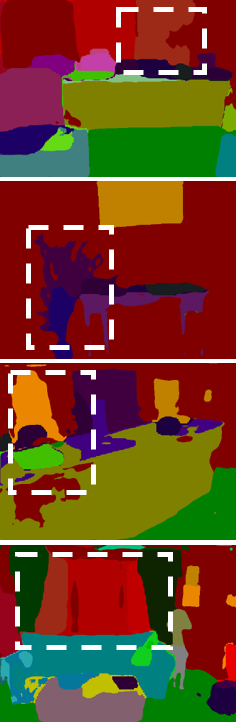}
		\caption{MetaFormer}
	\end{subfigure}
	\hspace{-0.7em}
	\begin{subfigure}[!t]{0.14\textwidth} 
		\includegraphics[width=0.95\linewidth]{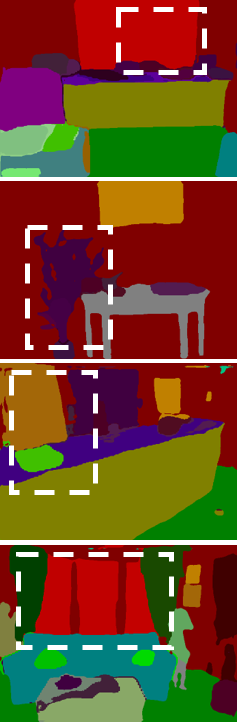}
		\caption{Ours}
	\end{subfigure}
        \hspace{-0.7em}
	\caption{Panoptic segmentation maps on NYUv2. Based on EMSAFormer, we compare our proposed model with models using different encoders. Our method withstands lighting variations and provides more complete segmentation of indoor plants and furniture.}
	\label{nyu}
\end{figure*}

We compare our model with existing methods on the NYUv2 dataset. Since the only available method for comparison in multi-task scene understanding is the one proposed by Fischedick et al. \cite{emsaformer}, we compared our model with semantic segmentation networks. As shown in Table \ref{tab：nyu}, benefiting from the depth information's representation of spatial structure, the RGB-D methods achieve higher accuracy than the RGB methods. Our method achieves the best semantic mIoU score of 49.82\% while performing multiple tasks simultaneously. Additionally, Figure \ref{nyu} shows a comparison of panoptic segmentation maps. Typically, outdoor lighting can interfere with indoor scene segmentation. However, our method successfully identifies the complete window outlines. Furthermore, the segmentation of indoor potted plants and furniture is also more complete.

\begin{table}[width=.85\linewidth,pos=!t]
\centering
\setlength{\tabcolsep}{25pt}
\renewcommand{\arraystretch}{1.1}
\caption{Comparison results on the NYUv2 dataset. We compare our proposed model with existing semantic segmentation models and list the input modalities and backbones of these models.}
\footnotesize
\begin{tabular}{lllc}
\toprule
Method     & Mode  &Backbone & Semantic mIoU$\uparrow$ \\ \midrule
MaskSup \cite{maskesup} & RGB & U-Net++ &39.31 \\
DIANet \cite{DualDIANet}& RGB &ResNet50 & 42.60 \\
MTML \cite{mtml} & RGB &ResNet50 & 41.51 \\
FDNet \cite{FDNet}& RGB & DenseNet264&47.40\\
PGT \cite{pgt} & RGB& Swin-S &46.43\\
SOSD-Net \cite{sosdnet}       & RGB   &DeepLabv3$+$ & 45.00         \\
DenseMTL \cite{densemtl}   & RGB   &ResNet101 & 40.84         \\
ZACN \cite{ZACN}& RGB-D &ResNet34&49.15\\
Malleable 2.5D \cite{malleable}&RGB-D&ResNet50&49.70 \\
ComPtr \cite{comptr} & RGB-D& Swin-T&49.20\\
MKE \cite{mke}        & RGB-D &ResNet101 & 48.88         \\
Link-RGBD \cite{linkrgbd} & RGB-D &ResNet50 & 49.50         \\
MMANet \cite{mmanet}     & RGB-D &R34-NBt1D & 49.62         \\
SGACNet \cite{sgacnet}    & RGB-D &R34-NBt1D & 49.40          \\
EMSAFormer \cite{emsaformer}& RGB-D &Swin v2 & 49.76         \\
Ours       & RGB-D &FasterNet-M & \textbf{49.82}         \\ \bottomrule
\end{tabular}
\label{tab：nyu}%
\end{table}%

\subsubsection{SUN RGB-D}
\begin{figure*}[!t]
	\centering
	\begin{subfigure}[!t]{0.14\textwidth}
		\includegraphics[width=0.95\linewidth]{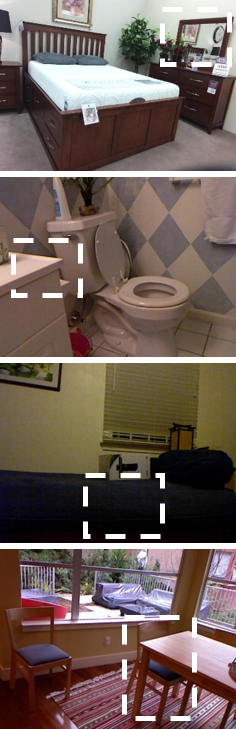}
		\caption{RGB}
	\end{subfigure}
	\hspace{-0.7em}
	\begin{subfigure}[!t]{0.14\textwidth} 
		\includegraphics[width=0.95\linewidth]{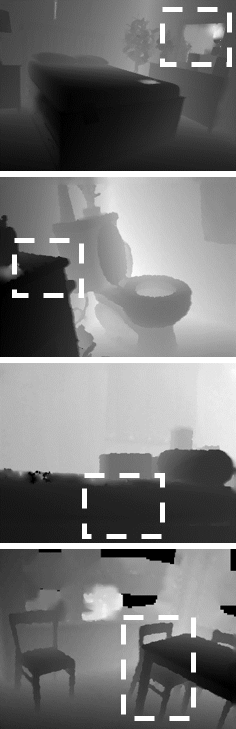}
		\caption{Depth}
	\end{subfigure}
	\hspace{-0.7em}
	\begin{subfigure}[!t]{0.14\textwidth} 
		\includegraphics[width=0.95\linewidth]{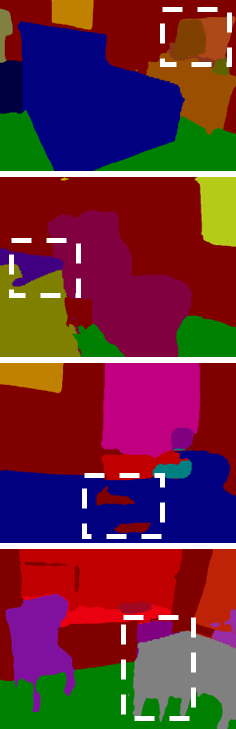}
		\caption{EMSAFormer}
	\end{subfigure}
	\hspace{-0.7em}
	\begin{subfigure}[!t]{0.14\textwidth}
		\includegraphics[width=0.95\linewidth]{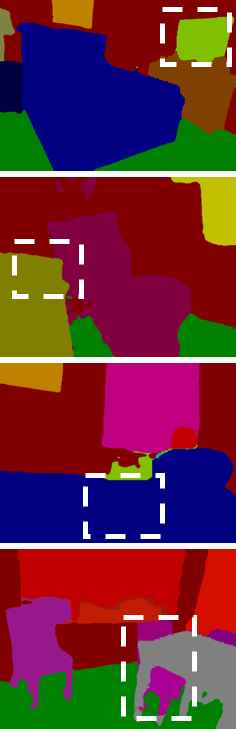}
		\caption{Ours}
	\end{subfigure}
	\caption{Panoptic segmentation maps on the SUN RGB-D dataset. We compare the panoptic segmentation maps of our proposed model with those of EMSAFormer. Our method accurately identifies and segments objects under low-lighting and occlusion conditions.}
	\label{sunrgbd}
\end{figure*}

\begin{table}[width=.85\linewidth,pos=!t]
\centering
\setlength{\tabcolsep}{25pt}
\renewcommand{\arraystretch}{1.1}
\caption{Comparison results on the SUN RGB-D dataset. We compare our proposed model with existing semantic segmentation models, using semantic mIoU as the evaluation metric.}
\footnotesize
\begin{tabular}{lllc}
\toprule
Method     & Mode  &Backbone& Semantic mIoU$\uparrow$ \\ \midrule
DIANet \cite{DualDIANet} & RGB &ResNet50 & 43.10 \\
CI-Net \cite{cinet}      & RGB   &ResNet101 & 44.30         \\
SSMA \cite{ssma}  & RGB   &ResNet50& 38.40         \\
SSMA \cite{ssma}  & RGB-D   &ResNet50& 43.90         \\
3DGNN \cite{3dgnn}  & RGB-D &VGG-Net & 40.20         \\
D-CNN \cite{dcnn2018}    & RGB-D &VGG-Net & 42.00         \\
EMSAFormer \cite{emsaformer}   & RGB-D &Swin v2 & 44.13          \\
Ours       & RGB-D &FasterNet-M & \textbf{45.56}         \\ \bottomrule
\end{tabular}
\label{tab：sunrgbd}%
\end{table}%

To assess the model's generalization performance across various scenes, we perform experiments on the SUN RGB-D dataset. As illustrated in Table \ref{tab：sunrgbd}, increasing the number of model layers allows CI-Net \cite{cinet} to mitigate the limitations caused by missing depth information in the RGB modality. In comparison, our approach surpasses others, achieving a semantic mIoU score of 45.56\%. Figure \ref{sunrgbd} showcases the panoptic segmentation maps. While reflections in the mirror can disrupt model perception, our model remains unaffected. In low-light conditions, it also provides more accurate segmentation of object contours. Additionally, in cases of object occlusion, our model effectively identifies the chair hidden behind the dining table.

\subsubsection{Cityscapes}
\begin{figure}[!t]
	\centering
	\includegraphics[width=4.5in]{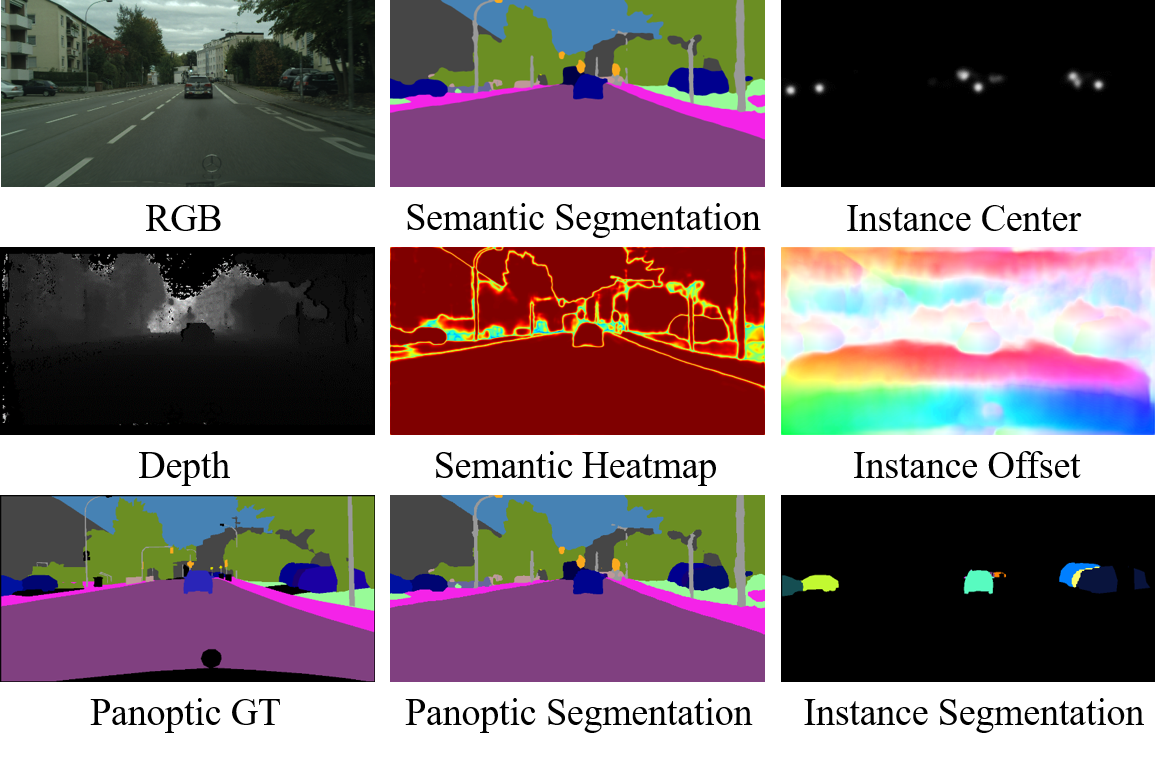}
	\caption{Visualization results on the Cityscapes dataset. We present the input information, ground truth (GT), the heatmap and segmentation map for semantic segmentation, as well as the instance center and offset for instance segmentation.}
	\label{cityscapes}
\end{figure}

\begin{table}[width=.85\linewidth,pos=!t]
\centering
\setlength{\tabcolsep}{25pt}
\renewcommand{\arraystretch}{1.1}
\caption{Comparison results on the Cityscapes dataset. Semantic mIoU is adopted as the evaluation metric.}
\footnotesize
\begin{tabular}{lllc}
\toprule
Method     & Mode  &Backbone& Semantic mIoU$\uparrow$ \\ \midrule
IkshanaNet \cite{ikshana}       & RGB &IkshanaNet & 54.40         \\
SegNet \cite{segnet} & RGB &VGG-Net & 56.10        \\
ENet \cite{Enet}   & RGB &ENet & 58.30         \\
Lovasz \cite{lovasz} & RGB   &ENet& 63.06         \\
ESPNet \cite{espnet}     & RGB   &ESPNet & 60.30         \\
EMSANet \cite{emsanet}  & RGB-D &ResNet34 NBt1D & 41.92          \\
EMSAFormer \cite{emsaformer}   & RGB-D &Swin v2 & 60.76          \\
Ours       & RGB-D &FasterNet-M & \textbf{65.11}         \\ \bottomrule
\end{tabular}
\label{tab：cityscapes}%
\end{table}%

Typically, indoor models perform less effectively when applied to outdoor scenes. The proposed model is primarily designed for indoor scenes. To further assess its adaptability, we conducted experiments on the outdoor Cityscapes dataset with the resolution set to 1024 $\times$ 512. In Fig. \ref{cityscapes}, we present visual results at various stages of instance segmentation, semantic segmentation, and panoptic segmentation. Even with blurry depth input, the model effectively segments vehicles, buildings, and streetlights. Additionally, the semantic heatmap clearly highlights the model's focus on key objects. Notably, the segmentation result is unaffected by the occlusion of the vehicle logo. Furthermore, we compare the proposed model with existing methods in Table \ref{tab：cityscapes}, showing superior performance with a semantic mIoU of 65.11\%.

\subsection{Limitation and discussion}
Although the proposed RGB-D scene understanding model has made significant strides in performance, efficiency, and task adaptability, it still faces certain limitations and areas that warrant further exploration. The enhanced fusion encoder leverages redundant feature information by sampling only a subset of features to improve processing speed. However, achieving a better balance between accuracy and processing time remains an open challenge. Future optimizations could employ Neural Architecture Search (NAS) to tailor the model for hardware constraints \cite{nas}. Another key challenge is the scalability of high-resolution inputs. As computational complexity increases, the current implementation struggles to process very high-resolution images or videos. Therefore, exploring more efficient architectures or compression techniques that can handle higher-resolution data without compromising performance is an important avenue for future research. Additionally, the model assumes that RGB-D inputs are well-calibrated and noise-free, which is not always the case in real-world applications. Consumer-grade depth sensors are often affected by issues such as reflections, transparent surfaces, and sparse measurements at object boundaries. Improving robustness to sensor noise and calibration errors will be a valuable direction for future work. In real-world scenarios, symbolic reasoning could augment data-driven models to handle logical constraints \cite{Reasoning}, such as ensuring that tables and chairs must be supported by the floor. Moreover, the current approach processes frames independently and disregards the temporal consistency of the data stream. Leveraging past predictions to improve current estimates in dynamic scenes, for instance, through memory networks or 3D convolutions \cite{memory}\cite{3Dconv}, could enhance consistency and reduce flickering in segmentation outputs.

Beyond these technical discussions, several ethical considerations must be addressed for practical applications. Privacy concerns arise when deploying RGB-D sensors in sensitive areas, as continuous scene understanding may inadvertently capture sensitive personal information. Data anonymization protocols and strict access controls should be implemented to comply with privacy regulations. Moreover, failure cases such as misclassification in cluttered or occluded environments can lead to incorrect robotic decisions, potentially causing safety risks. Systematic evaluation and transparent reporting of such failure modes are necessary to build trust and ensure reliability. Robotic environmental perception encompasses a wide range of scene features. While our work bridges important gaps between theoretical advances and practical applications, real-world deployment will require careful consideration of safety-critical failure modes, ethical constraints, and domain-specific adaptation challenges. Future research should prioritize not only algorithmic improvements but also comprehensive evaluations under diverse, realistic conditions to ensure reliable performance across different environments.

\section{Conclusion}\label{sec:Conclusion}
This paper presents an efficient RGB-D scene understanding model that integrates semantic segmentation, instance segmentation, orientation estimation, panoptic segmentation, and scene classification tasks into a unified framework. The fusion encoder effectively extracts complementary features from both RGB and depth data, addressing the feature redundancy issue. The introduction of the normalized focus channel layer and the context feature interaction layer further guides the model to focus on regions with high information content and spatial structures. Additionally, a multi-task adaptive loss function is designed, allowing the model to adjust its learning strategy in real time based on data variations, thus enabling more flexible joint training and performance optimization. Extensive experiments on the NYUv2, SUN RGB-D, and Cityscapes datasets demonstrate that the model achieves superior performance and faster speed.

Future research directions will focus on expanding the model's capabilities through the integration of additional modalities, including thermal imaging and point cloud data, to enhance robustness across diverse environmental conditions. We intend to investigate advanced optimization techniques, such as model pruning, quantization, and neural architecture search, to reduce computational latency and resource requirements. Additionally, we will explore data-efficient learning paradigms, including semi-supervised and few-shot learning approaches, to improve the model's applicability in situations with limited labeled data.

\bibliography{latest-cas-refs}
\end{document}